\pdfoutput=1 

\documentclass[twocolumn]{article}
\usepackage{adjustbox}
\usepackage{amstext}
\usepackage{amssymb} 
\usepackage[greek, british]{babel}
\usepackage{booktabs}
\usepackage{csquotes} 
\usepackage{setspace}
\usepackage{siunitx}
\usepackage[caption=false]{subfig}
\usepackage{textcomp}
\usepackage{tikz}
\usepackage{varioref}

\definecolor{fhg}{RGB}{23,156,125}
\usepackage[colorlinks=true, pdfstartview=FitV, allcolors=fhg, pagebackref=true]{hyperref}

\usetikzlibrary{intersections,3d,decorations.text,shapes.arrows,positioning,fit,backgrounds,trees,positioning,fit,arrows,decorations.pathreplacing, calc,math,through,arrows.meta }
\definecolor{tikzBoxColourLightGreen}{RGB}{192,224,0}
\definecolor{tikzBoxColourOrange}{RGB}{224,128,64}
\definecolor{tikzBoxColourBlue}{RGB}{128,160,192}
\newcommand{\tikzBoxSize}{0.15cm}
\DeclareRobustCommand\tikzBoxLightGreen{\tikz\node[rectangle,fill=tikzBoxColourLightGreen,minimum width=\tikzBoxSize,minimum height = \tikzBoxSize,] (r) at (0,0) {};}
\DeclareRobustCommand\tikzBoxOrange{\tikz\node[rectangle,fill=tikzBoxColourOrange,minimum width=\tikzBoxSize,minimum height = \tikzBoxSize,] (r) at (0,0) {};}
\DeclareRobustCommand\tikzBoxBlue{\tikz\node[rectangle,fill=tikzBoxColourBlue,minimum width=\tikzBoxSize,minimum height = \tikzBoxSize,] (r) at (0,0) {};}

\title{An annotated instance segmentation XXL-CT data-set from a historic airplane}
\author{
	\href{https://orcid.org/0000-0002-3429-732X}{\includegraphics[scale=0.06]{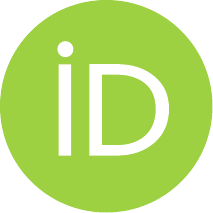}}Roland Gruber$^{1,2}$, 
	Nils Reims$^{1}$,
	Andreas Hempfer$^{3}$, 
	\href{https://orcid.org/0000-0002-5590-2920}{\includegraphics[scale=0.06]{./data/full/orcid.pdf}}Stefan Gerth$^{1}$, \\
	Michael Böhnel$^{1}$, 
	Theobald Fuchs$^{1}$, 
	Michael Salamon$^{1}$,
	\href{https://orcid.org/0000-0003-0840-8695}{\includegraphics[scale=0.06]{./data/full/orcid.pdf}}Thomas Wittenberg$^{1,2}$\\ 
	\\
	$^{1}$ \small Fraunhofer IIS, Fraunhofer Institute for Integrated Circuits IIS, Fürth, Germany \\
	$^{2}$ \small Friedrich-Alexander-Universität Erlangen-Nürnberg, Germany \\
	$^{3}$ \small Deutsches Museum, München, Germany \\
}

\date{2024-02-05}

\begin{document}
	
	\maketitle	
	\begin{abstract}	
		The Me\,163 was a Second World War fighter airplane and a result of the German air force secret developments. One of these airplanes is currently owned and displayed in the historic aircraft exhibition of the \enquote{Deutsches Museum} in Munich, Germany. To gain insights with respect to its history, design and state of preservation, a complete CT scan was obtained using an industrial XXL-computer tomography scanner.
		
		Using the CT data from the Me\,163, all its details can visually be examined at various levels, ranging from the complete hull down to single sprockets and rivets. However, while a trained human observer can identify and interpret the volumetric data with all its parts and connections, a virtual dissection of the airplane and all its different parts would be quite desirable. Nevertheless, this means, that an instance segmentation of all components and objects of interest into disjoint entities from the CT data is necessary. 		
		
		As of currently, no adequate computer-assisted tools for automated or semi-automated segmentation of such XXL-airplane data are available, in a first step, an {\em interactive} data annotation and object labelling process has been established. So far, seven $512 \times 512 \times 512$ voxel sub-volumes from the Me\,163 airplane have been annotated and labelled, whose results can potentially be used for various new applications in the field of digital heritage, non-destructive testing, or machine-learning. These seven annotated and labelled data sets are available from  \cite{fraunhoferMe163InstanceSegmentationDataset}.  
		
		This work describes the data acquisition process of the airplane using an industrial XXL-CT scanner, outlines the interactive segmentation and labelling scheme to annotate sub-volumes of the airplane's CT data, describes and discusses various challenges with respect to interpreting and handling the annotated and labelled data.
	\end{abstract}
	
	\section{Introduction}			
\begin{figure*}
		\centering{}
		\includegraphics[width=1.0\textwidth, keepaspectratio]{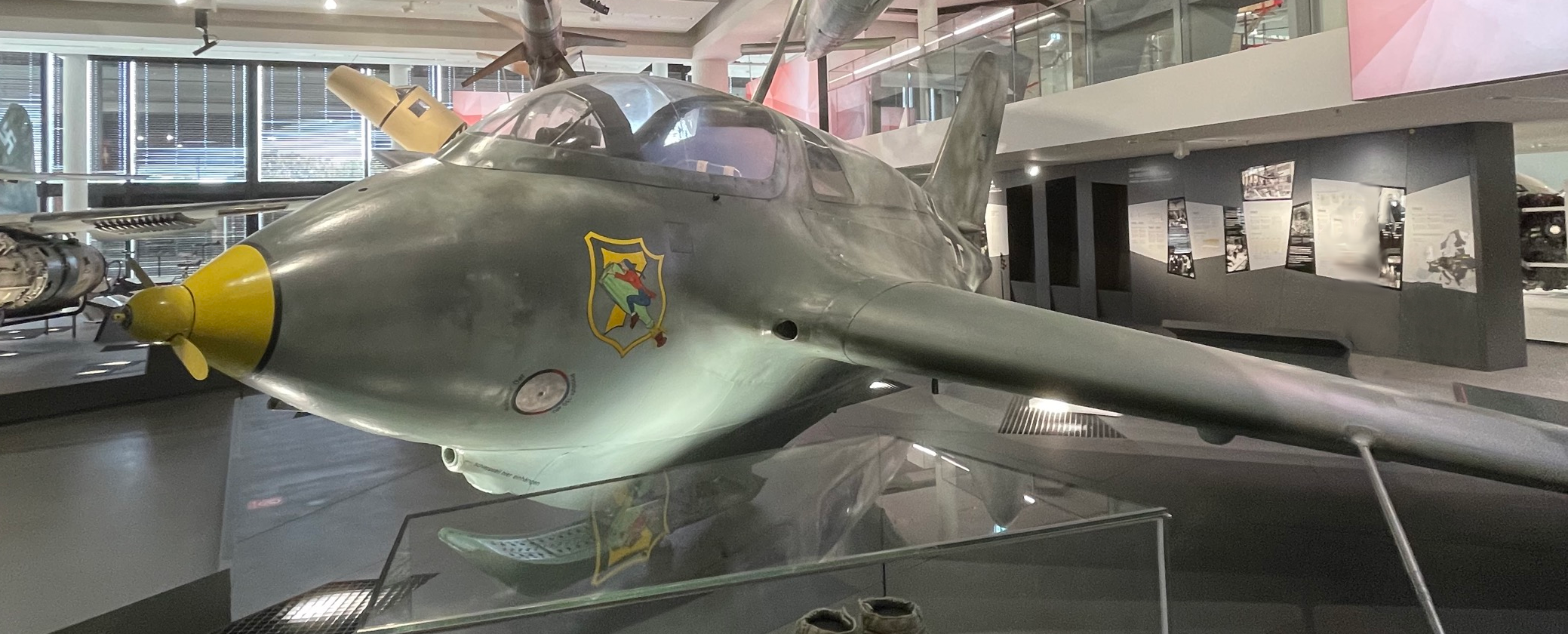}	
		\caption{\label{fig:data-ME163-full} Image of the Messerschmitt Me\,163 in the historic aircraft exhibition of the \enquote{Deutsches  Museum}}
	\end{figure*}

	The Messerschmitt Me\,163, see Figure \ref{fig:data-ME163-full}, was a German fighter airplane with a rocket engine during the Second World War, and was part of the secret developments of the German air force  \cite{Cammann2001-1}. With its unique rocket engine, it was the first piloted aircraft to reach a maximum speed of about \SI{1000}{\kilo\metre\per\hour}. Of the 350\,Me\,163s built between 1941 and 1945, only ten examples survive in museums, one of which is displayed in the in the historic aircraft exhibition of the \enquote{Deutsches Museum} in Munich, Germany. 
	
	To gain new insights into the history, design and state of preservation of this unique and historical airplane, a complete CT scan was obtained using the XXL-computer tomography scanner of the Fraunhofer IIS' development center for X-ray technologies EZRT in Fürth, Germany \cite{Salamon2016}, see Figure \ref{fig:data-me163}. 
	
	Besides viewing and examining the XXL-CT data in detail using adequate interactive 3D-volume reader and viewer software \cite{Hadwiger2008} also, the airplane’s many individual components are of interest, such as the screws, wheels, sprockets, rivets, and much more. 
	
	To obtain more information about these parts, their distribution within the airplane and their spatial and functional relationships to each other ideally an automated, semi-automatic, or purely manual instance segmentation or partitioning of all components  and objects of interest into disjoint parts from the CT data is necessary.		
	
	To this end, different automated CT volume-seg\-mentation methods of different complexity could be applied to obtain a set of segmented airplane parts. Nevertheless, all automatic as well as semi-automatic 3D image segmentation methods usually depend strongly on the availability of sufficient and adequate labelled reference data needed for the parameter optimisation or parameter training, as well as the sufficient evaluation of the developed delineation approaches. As the Me\,163 aeroplane is an example of a unique object with partially very exceptional and matchless components, of which only one CT-scan exits (also known as \emph{lot-one-problem}), adequate automatic or semi-automatic segmentation methods are currently not available, enabling and supporting the  delineation of the airplane's different components. 
	
	
	However, if -- as a first step -- some adequate labelled reference data from such an XXL-CT airplane scan would be available, the development of new segmentation methods, either based on traditional image processing methods or alternatively using novel deep-learning approaches (e.g.\ employing deep convolutional neural networks (DCNNs) \cite{Gruber2020, 2021-Du-IEEE, Hafiz2020}, could be developed and evaluated more efficiently. Especially as the performance of such DCNNs on vision tasks tends to increase logarithmically based on the volume of training data \cite{x2}. Thus,  in order to optimize an automation segmentation scheme, a large set of well-curated ground truth data sets is of most importance \cite{2-1-72-Deng2009, 2-1-72-Lin2014b}. 		    
	
	Hence, within this contribution, we will provide historical background to the Me\,163 and its current stay in the ‘Deutsches Museum’ (Section \ref{sec:the-me-163}), describe the data acquisition process using an XXL CT scanner  (Section \ref{sec:data-acquisition}), outline the interactive labelling and annotation process of some distinct sub-volumes of the airplane (Section \ref{sec:data-annotation}), and discuss various challenges with respect to interpreting and handling the annotated and labelled data (Section \ref{sec:challenges}). Furthermore, we introduce a matrix-based metric to compare two (manually or automatic) labelled segmentations (Section \ref{sec:metric}) which can handle erroneous split or merged segments as well as voxel overlap of the segments.
	
	Seven of the sub-volumes together with their manually obtained annotations shall be made public available \cite{fraunhoferMe163InstanceSegmentationDataset} to be used in the future by researchers in the fields of digital heritage, non-destructive testing, machine vision and/or artificial intelligence to visualize and interact with as well as to develop, train, optimise and evaluate novel 3D-instance segmentation approaches.  
	
	\section{The Me\texorpdfstring{\,}{ }163} \label{sec:the-me-163}
	The Messerschmitt Me\,163 \cite{brown19773wings, butlar1994, Cammann2001-1} in the historic aircraft collection and exhibition of the \enquote{Deutsches Museum} (see Figure \ref{fig:data-ME163-full}) is still a mysterious plane. The British' Royal Air Force (RAF) gifted it to the museum in 1964, but since the ID plate in the nose is empty, not much is known about its operational history in the Second World War or its second life in Great Britain. After it was captured in 1945, the plane was modified for flight-testing by the RAF. When an accident with another Me 163 nearly killed a test pilot, the Me 163s were kept only as technological curiosities. Some were scrapped after, some found their way into museums around the globe.
	
	The British had realized that this alleged Nazi \enquote{wonder weapon} was more of a danger for its pilots rather than allied planes. Developed from innovative tailless gliders by Alexander Lippisch and fitted with a Walter HWK\,109-509 rocket engine with \SI{14.7}{\kN} of thrust in 1941, the Me\,163 reached exceptional speeds and climb rates. The small and light airframe with its thick, swept wings reached Mach 0.84 and could climb up to \SI{81}{\metre\per\second}. These achievements, however, came at a very high price: With no space for a retractable landing gear, the wheels were jettisoned after takeoff, often bouncing back and damaging the plane. The rocket fuel was depleted in just seven minutes, leaving very little time to reach the enemy. The armament was weak and unsuited for the purpose of the Me\,163, that is intercepting heavy allied bombers. When gliding back to base, pilots could evade attacking fighters thanks to the good maneuverability of the Me\,163, only to be sitting ducks after they came to a halt on the landing skid. What really made the plane an unacceptable hazard for pilots and ground crew was its highly flammable rocket fuel: \enquote{C-Stoff} and \enquote{T-Stoff} (the latter 80\,\% hydrogen peroxide) exploded on contact and fumes could dissolve any organic matter. Fatal accidents at take-off or landing were common.
	
	The Me\,163, as well as the so-called weapons \enquote{V1} and \enquote{V2}, embodies a widespread belief in innovative technology as a miraculous savior from vastly superior allied air power. Forced laborers, willingly exploited by the German industry by the hundreds of thousands, had to build many parts of the plane in murderous conditions. In the end, the approx. 350\,Me\,163s produced in total, shot down only nine heavy allied bombers between 1943 and 1945. In telling us about the hubris of its engineers as well as cultural aspects of technology, the Me\,163 is a highly sought-after study object.
	
	\section{Data acquisition} \label{sec:data-acquisition}
\begin{figure}
		\centering{}
		\subfloat[\label{fig:data-me163-body}]
		{\includegraphics[width=0.45\columnwidth,height=0.5\textheight,keepaspectratio]
			{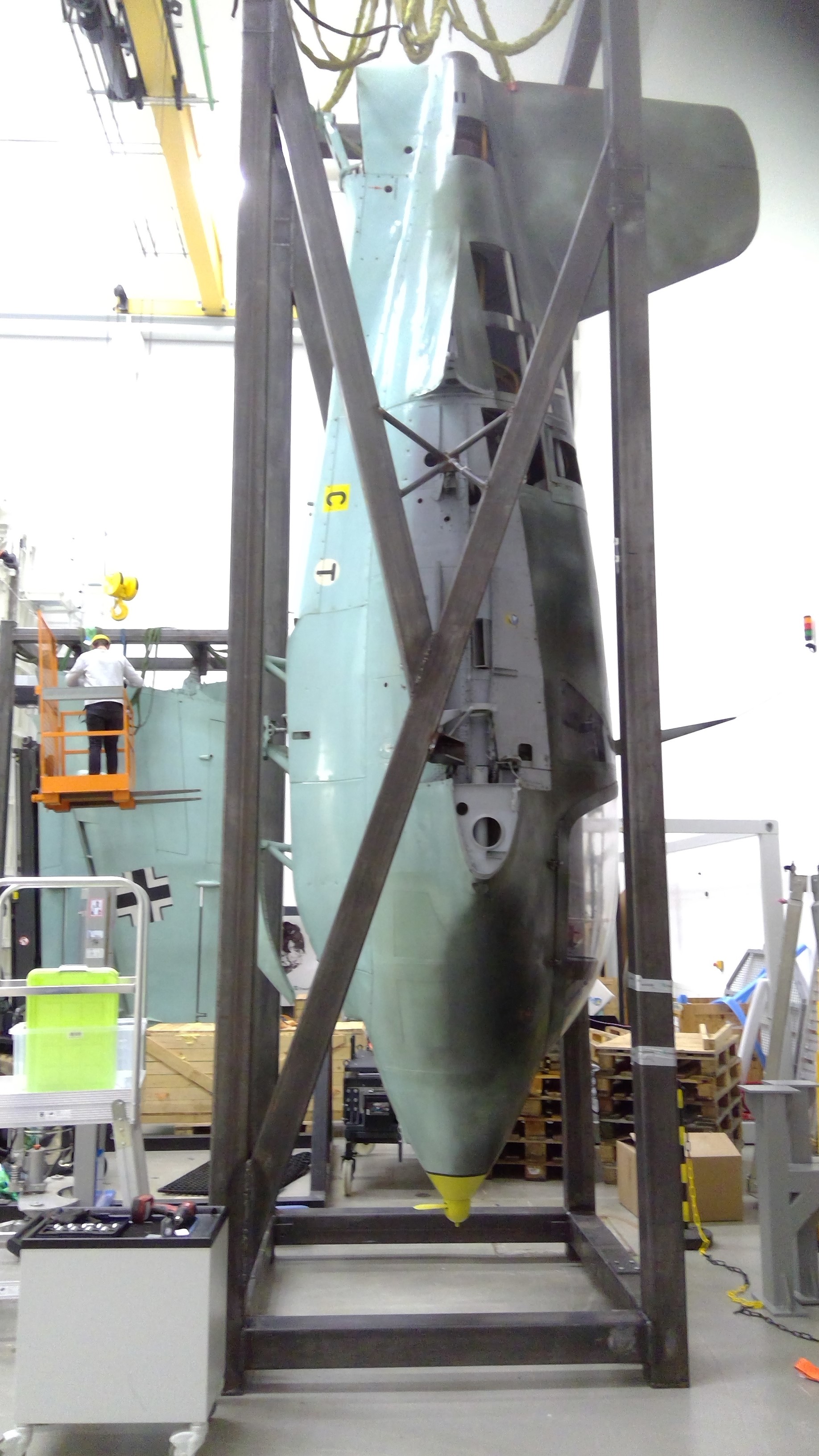}}
		~\subfloat[\label{fig:data-me163-wings}]
		{\includegraphics[width=0.45\columnwidth,height=0.5\textheight,keepaspectratio]
			{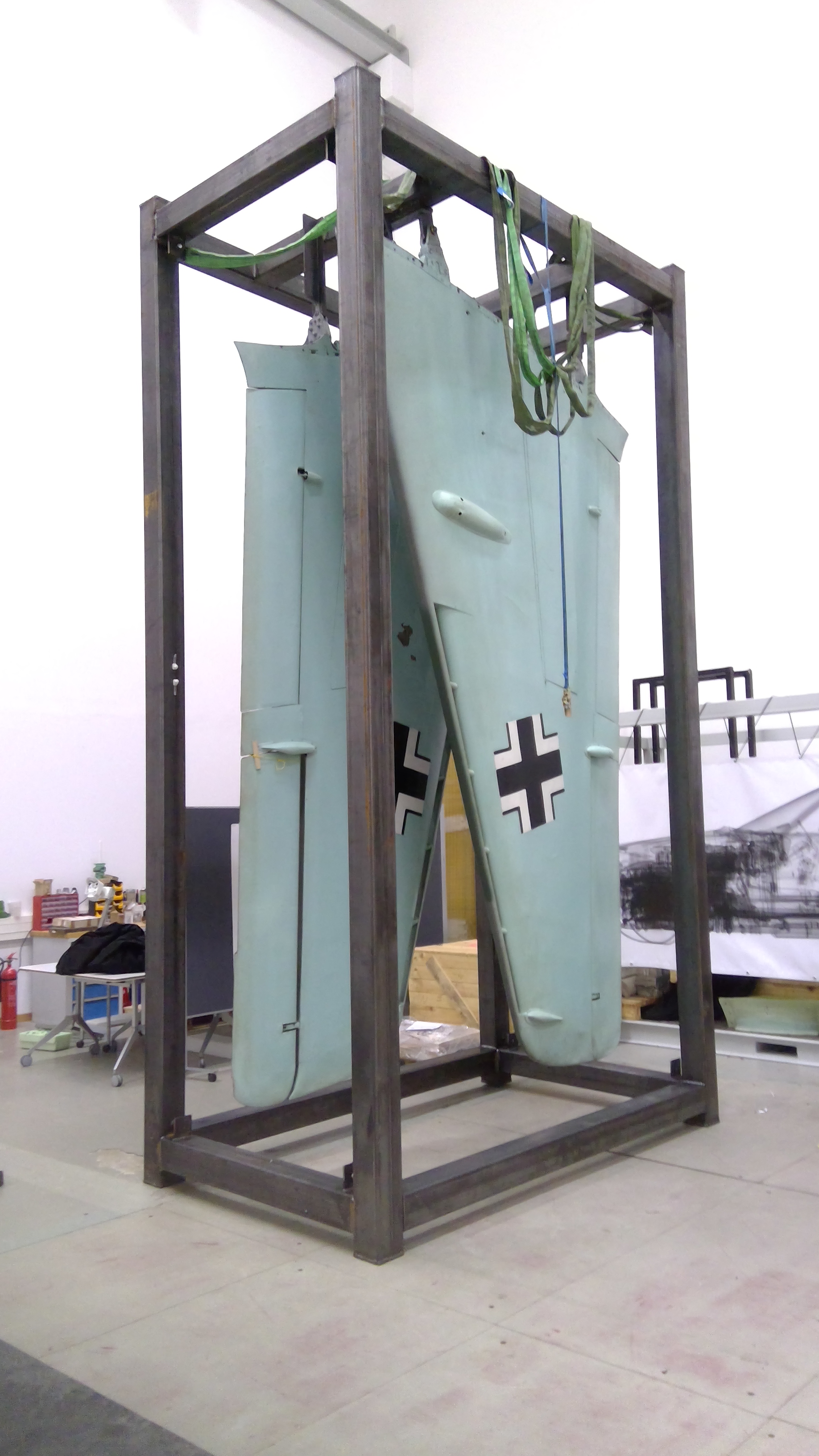}}	
		\caption{\label{fig:data-me163}Fuselage (Figure \ref{fig:data-me163-body}) and wings (Figure \ref{fig:data-me163-wings}) of the Me\,163 airplane inside the mounting brackets for the CT scan.}
	\end{figure}

\begin{figure*}
		\centering{}
		\subfloat[\label{fig:data-me163-reko-fuselage}]
		{\includegraphics[height=0.35\textheight,keepaspectratio]
			{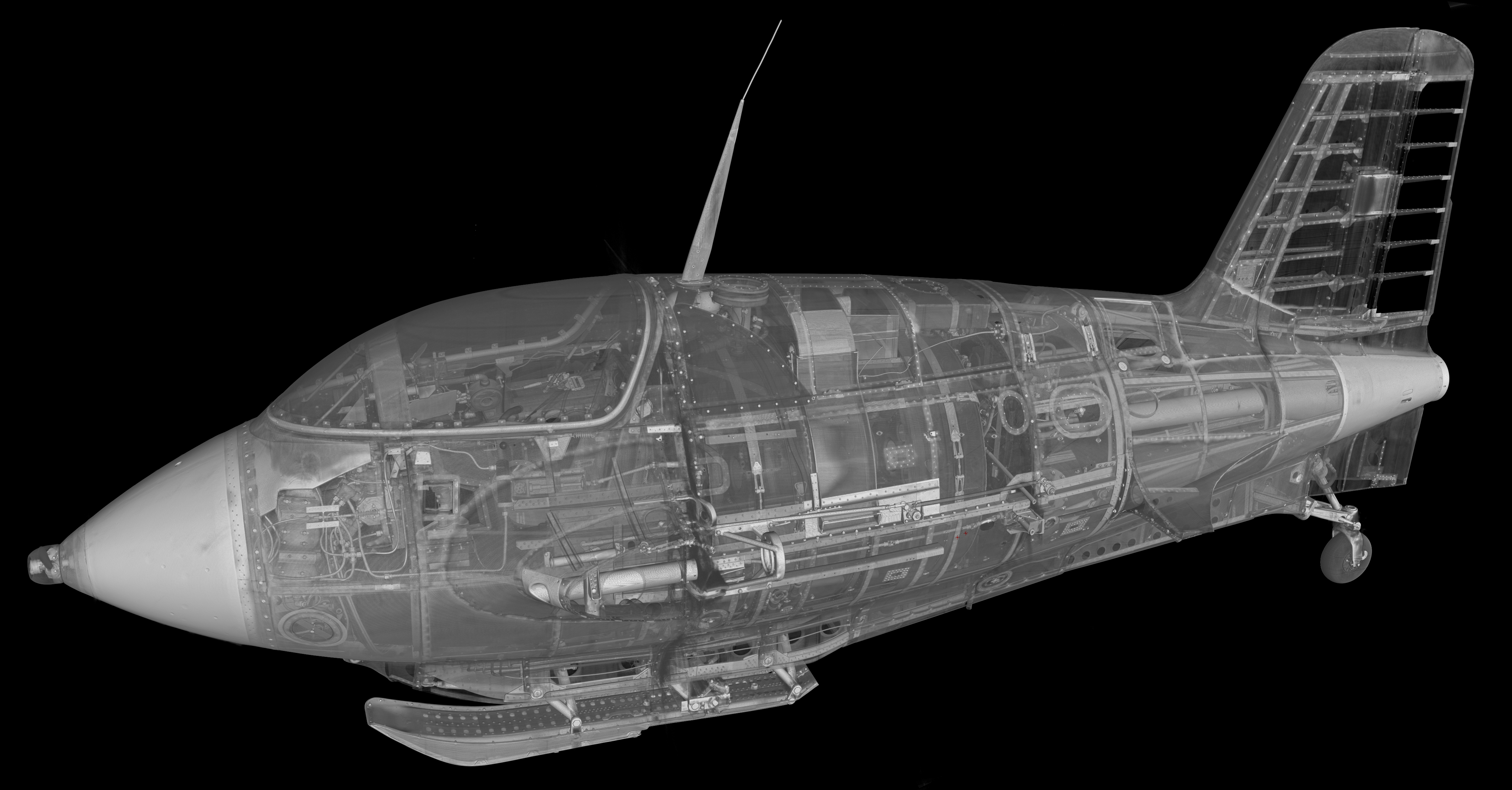}}	
		~\subfloat[\label{fig:data-me163-reko-roi}]
		{\includegraphics[height=0.35\textheight, interpolate=false,keepaspectratio]
			{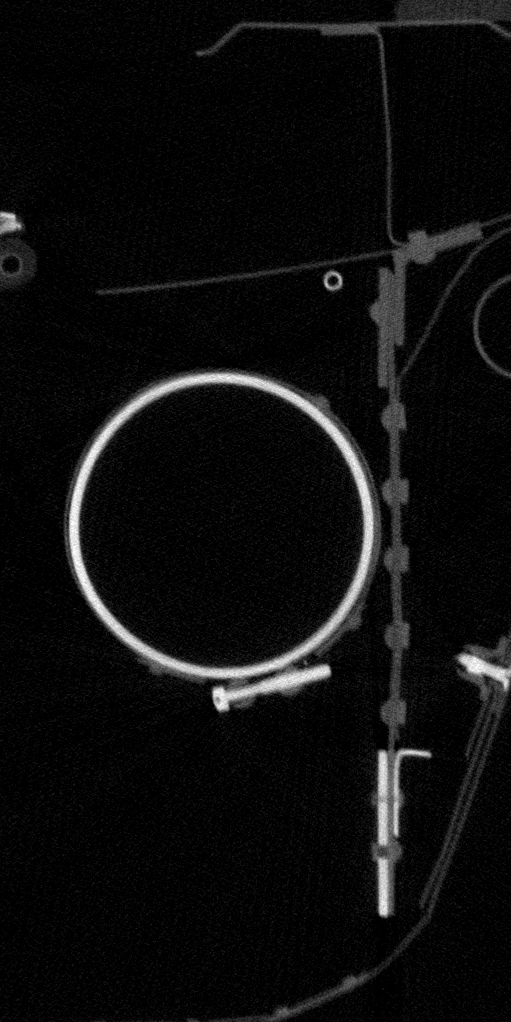}}
		\caption{\label{fig:data-me163-reko} Rendering of the reconstructed fuselage of the scanned airplane (Figure \ref{fig:data-me163-reko-fuselage}) and detail (Figure \ref{fig:data-me163-reko-roi}) located at approximately the midpoint of the fuselage between the nose and the tail of the airplane.}
	\end{figure*}

	The XXL-CT data-set of the historic Me\,163 airplane was acquired at the XXL-CT facility of the Fraunhofer IIS' development center for X-ray technologies EZRT in Fürth, Germany \cite{Salamon2016}. To cover the complete airplane, four subsequent CT-scans were performed, two for the fuselage (see Figure \ref{fig:data-me163-body}) and two for the disassembled wings (see Figure \ref{fig:data-me163-wings}). Afterwards, the two sub-data-sets for the fuselage and the two sub-data-sets for the wings were each manually merged into one data-set (see Figure \ref{fig:data-me163-reko-fuselage} left).
	
	In total the four CT scans of the airplane parts needed approximately 17\,days to complete. To provide enough performance to permeate the airplane with X-rays, a linear accelerator X-ray source with \SI{9}{\mega\electronvolt} was used. The distance between the X-ray source and the detector was set to $d_{\textrm{S-D}}  =$ \SI{12}{\metre} and the source-to-object distance to about $d_{\textrm{S-O}} = $ \SI{10}{\metre}. 
	
	The use of a line detector with a width of $w = $ \SI{4}{\metre} and a pixel spacing of \SI{400}{\micro\metre} results in a horizontal resolution of 9,984\,pixels. Using a vertical stepping motor, projective raw images with a spatial resolution of $9984\times5286$\,pixels are obtained. The magnification of 1.2 leads to a horizontal voxel resolution of \SI{330} \times\,\SI{330}{\micro\square\metre} and a vertical sampling of \SI{600}{\micro\metre} within the reconstructed volume.
	The obtained 16 bit data volumes of the two fuselage parts (see Figure \ref{fig:data-me163-reko}) have file sizes of $6,144\times9,600\times5,288$\,voxels \--  or approximately \SI{609}{\giga\byte} for the front part \-- and $6,144\times9,600\times5,186$\,voxels \-- or \SI{567}{\giga\byte} \-- for the rear part of the hull.			
	
	As expected, and can be seen in Figure \ref{fig:data-me163-reko-roi} as well as the last column of Table~\ref{tab:description-segmentation}, most of the airplane's reconstructed interior consists of empty space or air. Apart from that, the CT volumes depict mainly a plethora of thin metal sheets, which have poorly or barely visible edge transitions to the adjacent metal sheets.
	
	For the many cases where two metal sheets butt together, semantic information must be used to decide on the correct object boundaries between the entities. In addition, many regions in the XXL-CT volume are severely affected by artefacts from the data acquisition and reconstruction such as beam hardening or scattered radiation, especially in the vicinity of solid thick walled metal structures.	
	
	\section{Data Annotation} \label{sec:data-annotation}		
	Even though manual data labelling is currently referred to as the ‘gold standard’ for unique complex image data \cite{Wissler2014}, the required resources are quite high with respect to experienced staff and delineation time, even if specialized annotation pipelines (e.g. \cite{x59} and \cite{Gonda2021} ) allowing image processing guided annotation, proofreading of inference results and model refinements, are applied for this task. 
	
	Hence, to reduce the costs of experts needed for manual or interactive image labelling tasks, so-known ‘crowd-sourcing’ approaches have been proposed and partially established \cite{2-1-72-Deng2014, Romanowski-2019, Chen2016UsingCF}. 	
	
	Nevertheless, to be effective, crowd-sourcing also profits strongly from specialized data management, annotation tool and soft skills of the annotators \cite{16-18-58}. However, besides the huge amount of organizational, legal and logistic overhead, one drawback of crowd-sourcing is the limited understanding of the annotators about the annotation problem at hand and the complexity of the complex 3D data depicting the various objects. 				
	
	To somehow make a compromise between experts and crowd-sourcing, each individual sub-volume was initially annotated and labelled by an first annotator, and the thus acquired annotation was subsequently proofread and corrected by a second experienced annotator. 
	
	The complete annotation of the first two 512\textthreesuperior \,sub-volumes each needed about 350\,working hours (or approximately two months with 40\,hour per weeks), as the first annotator was trained on these sub-volumes and they contained many segments compared to later more empty sub-volumes. The manual annotation of each of these subsequent sub-volumes took about 10\,\% to 50\,\% of that time, mostly depending on the sub-volume complexity. The subsequent correction by different but trained annotators took about the same amount of time, or 4 to 120\, hours per sub-volume.
	
	The following Section \ref{subsec:description-of-data} will give a brief overview of the annotated XXL-data, while the used annotation pipeline will be introduced in Section \ref{sec:annotation-pipeline}. 
	
	\subsection{Description of data} \label{subsec:description-of-data}
	\begin{table*}
		\begin{adjustbox}{center}
			\begin{tabular}{ccrrrrcr} \toprule
				Sub -  	& coordinates 		& \# segments		& Minimum 		& Maximum 		 	& Median 			& foreground\\ 
				volume 	& in		  		& (prior post- 		& {segment size}& {segment size} 	& {segment size}   	& voxel  	\\ 
				& XXL-Volume		& processing)		& [voxel] 		& [voxel] 		 	& [voxel] 			& [\%]		\\ \midrule
				
				$V_1$ 	& (3072, 4608, 0)	& 5 (5) 			& 2,468 		& 1,210,717 		& 47,849 			& 1.3 		\\ 
				$V_2$ 	& (3072, 5120, 0) 	& 7 (7) 			& 1,277 		& 1,173,579			& \bf{78,659}		& 1.1 		\\ 
				$V_3$ 	& (3072, 5632, 0) 	& 14 (13)			& 1,147 		& 889,731			& 9,790 			& 1.3 		\\ 
				$V_4$ 	& (3072, 6144, 0) 	& 33 (33) 			& 1,293			& \bf{1,768,078}	& 2,217				& 4.5 		\\ 
				$V_5$ 	& (3072, 6656, 0) 	& {\bf 169 (159)} 	& 187	 		& 1,545,773			& 3,853				& \bf{9.4}	\\ 
				$V_6$ 	& (3072, 7168, 0) 	& 108 (94)			& \bf{158}		& 1,567,273			& 4,039				& 5.6 		\\ 
				$V_7$ 	& (3072, 7680, 0) 	& 9 (9) 			& 509 			& 73,302			& 1,209				& 0.1 		\\ \bottomrule
			\end{tabular}		
		\end{adjustbox}
		\caption{\label{tab:description-segmentation} Key metrics of the annotated sub-volumes depicted in Figure \ref{fig:description-3D}.}
	\end{table*}
	
\begin{figure*}
		\centering{}
		\includegraphics[width=1.0\textwidth,keepaspectratio]{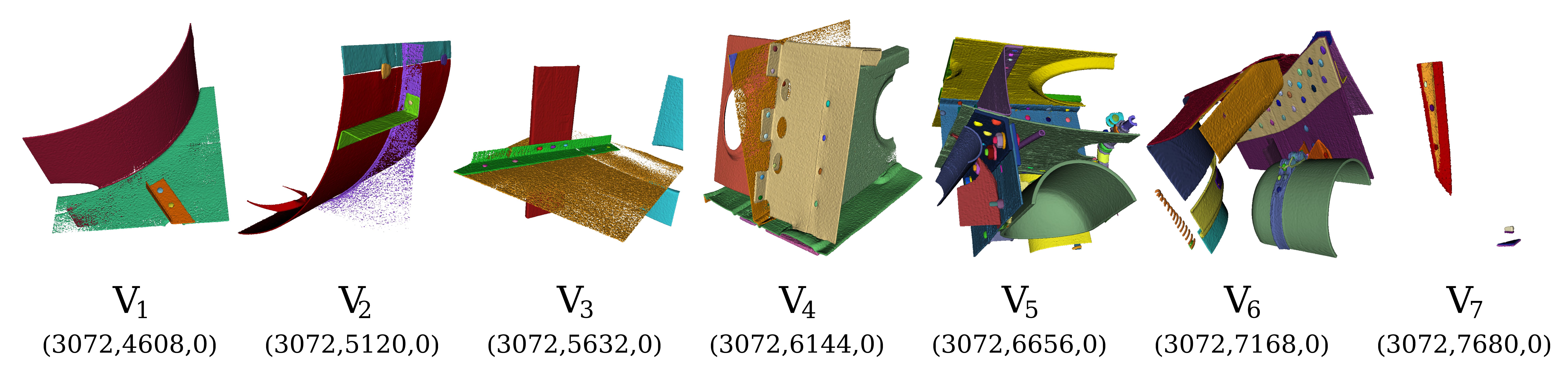}
		\caption{\label{fig:description-3D}Examples of 3D-renderings of manually annotated and labelled sub-volumes from the XXL-Scan of the Me\,163, depicting various semantic objects of different types, shapes, and materials.}
	\end{figure*}

	Figure \ref{fig:description-3D} provides examples of  3D-renderings of annotated and labelled sub-volumes. Each sub-volume contains between 5 and 172 individual object entities of various sizes, materials, and types. 
	
	Table~\ref{tab:description-segmentation} gives a brief overview of the depicted objects in the annotated sub-volumes. Even though the regarded sub-volumes are located in the centre of the airplane (see Figure~\ref{fig:data-me163-reko-fuselage}), the 0-coordinates in the second column indicate that they are placed exactly at the border between the two sub-scans. It can be seen, that the largest object in the Table \ref{tab:description-segmentation}, being a complex metal sheet, consisting of 1,768,078\,voxels (with an equivalent of approximately \SI{116}{\cubic\centi\metre}), while the smallest object being a rivet contains only of 158\,voxels (with an equivalent of \SI{0.1}{\cubic\centi\metre}). Both of these objects are bounded by their respective side surfaces of their surrounding sub-volumes and actually extend beyond them into adjacent sub-volumes.
	
	Overall annotated sub-volumes, approximately 93\,\% of all voxels refer to background data, namely air, while only 7\,\% (or 62.6\,million voxels) relate to data of the depicted objects. These comprise a sum of 344\,segments. 
	
\begin{figure*}
		\begin{centering}
			\subfloat[\label{fig:description-example-volume}]
			{\includegraphics[width=0.33\textwidth,height=0.25\textheight,keepaspectratio]
				{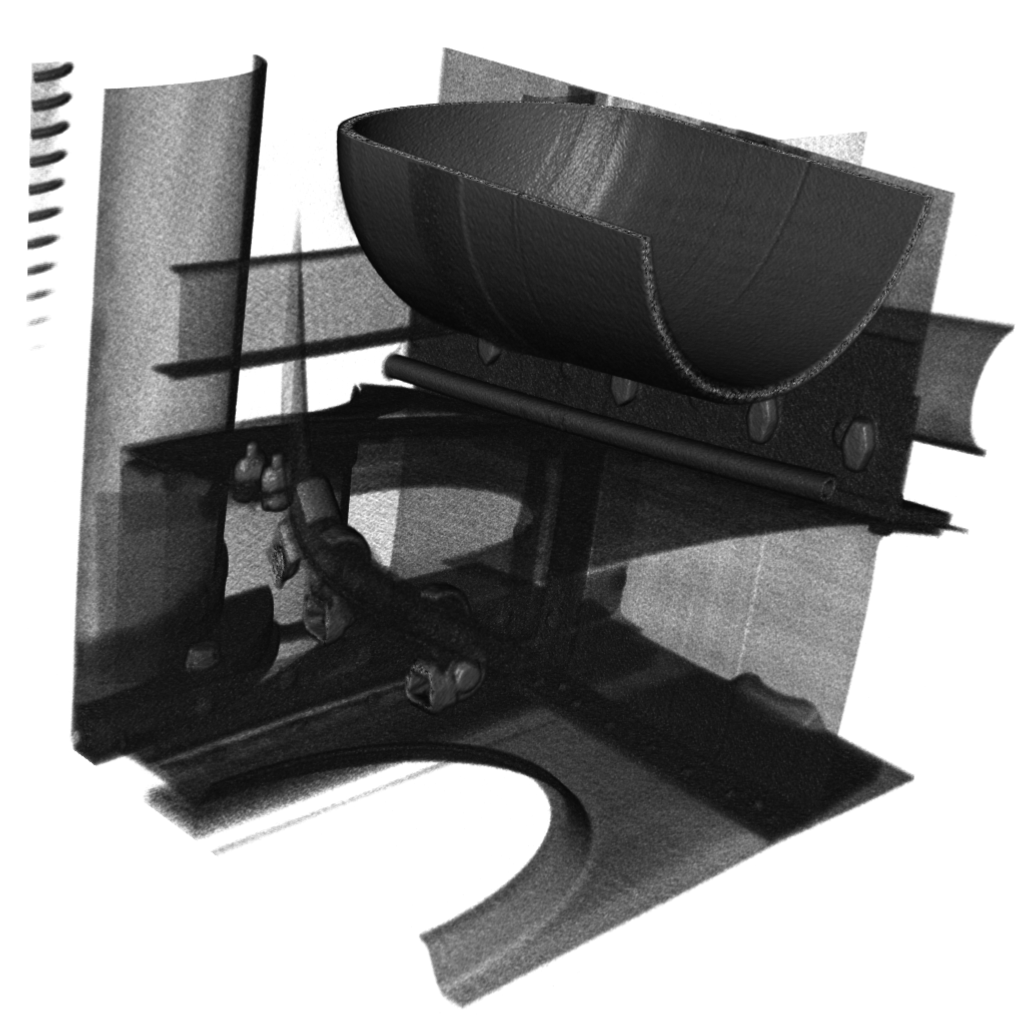}}	
			~\subfloat[\label{fig:description-example-all}]
			{\includegraphics[width=0.33\textwidth,height=0.25\textheight,keepaspectratio]
				{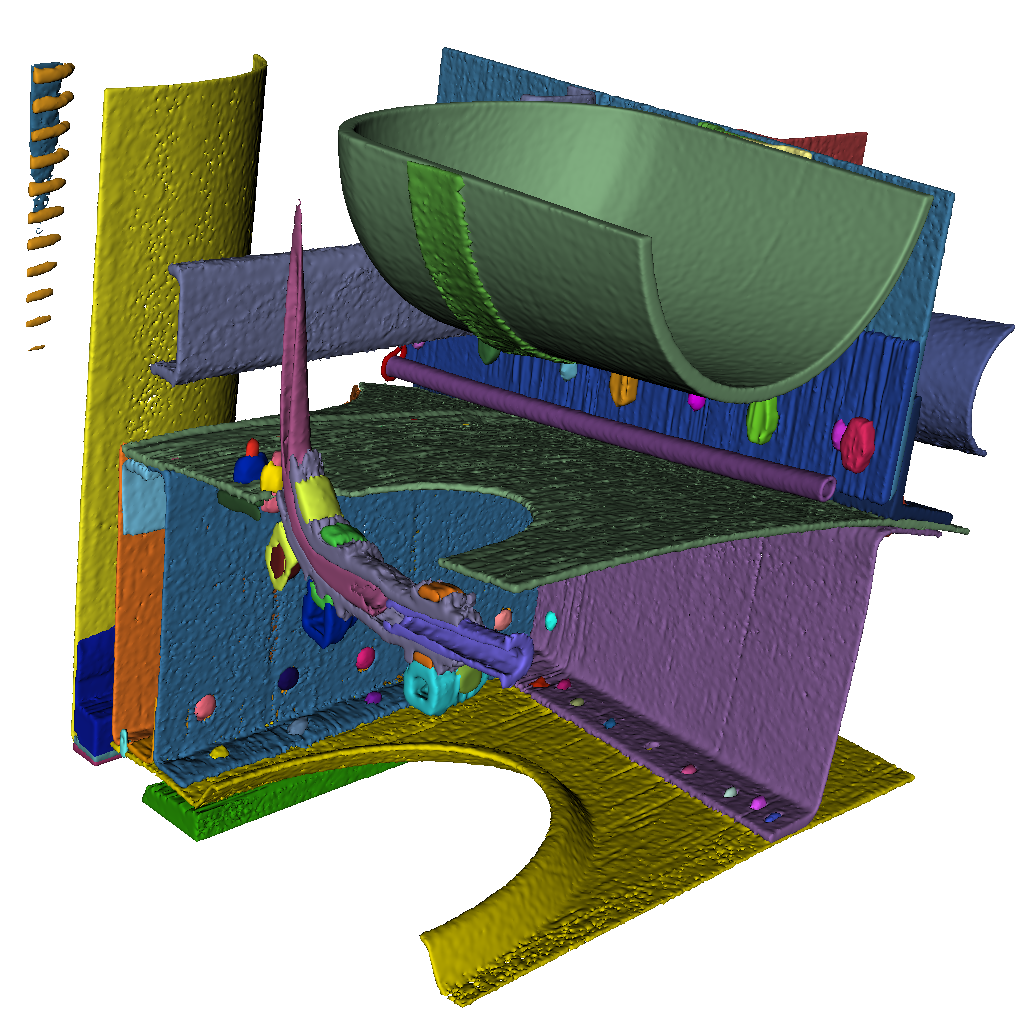}}
			~\subfloat[\label{fig:description-example-metallSheet}]
			{\includegraphics[width=0.33\textwidth,height=0.25\textheight,keepaspectratio]
				{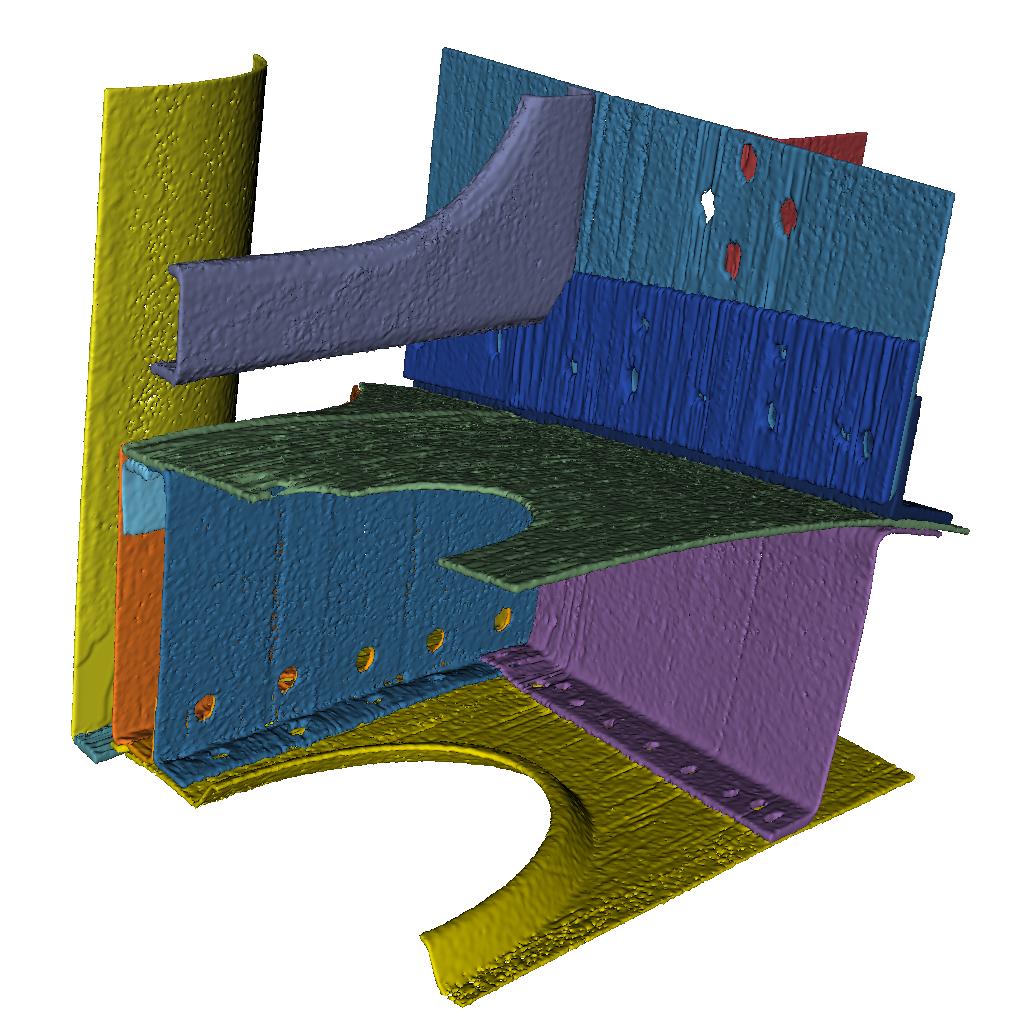}}
			
			\subfloat[\label{fig:description-example-pipes}]
			{\includegraphics[width=0.33\textwidth,height=0.25\textheight,keepaspectratio]
				{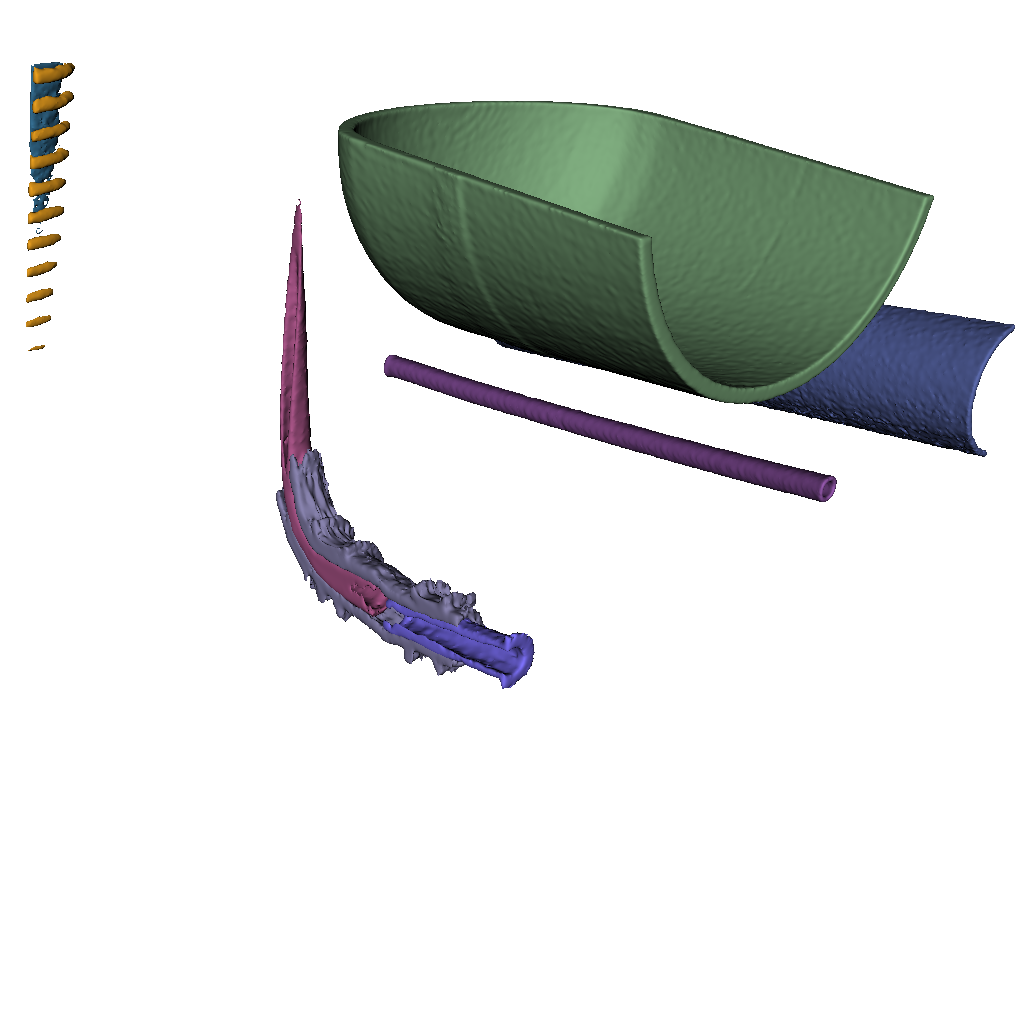}}	
			~\subfloat[\label{fig:description-example-rivetsAndBolts}]
			{\includegraphics[width=0.33\textwidth,height=0.25\textheight,keepaspectratio]
				{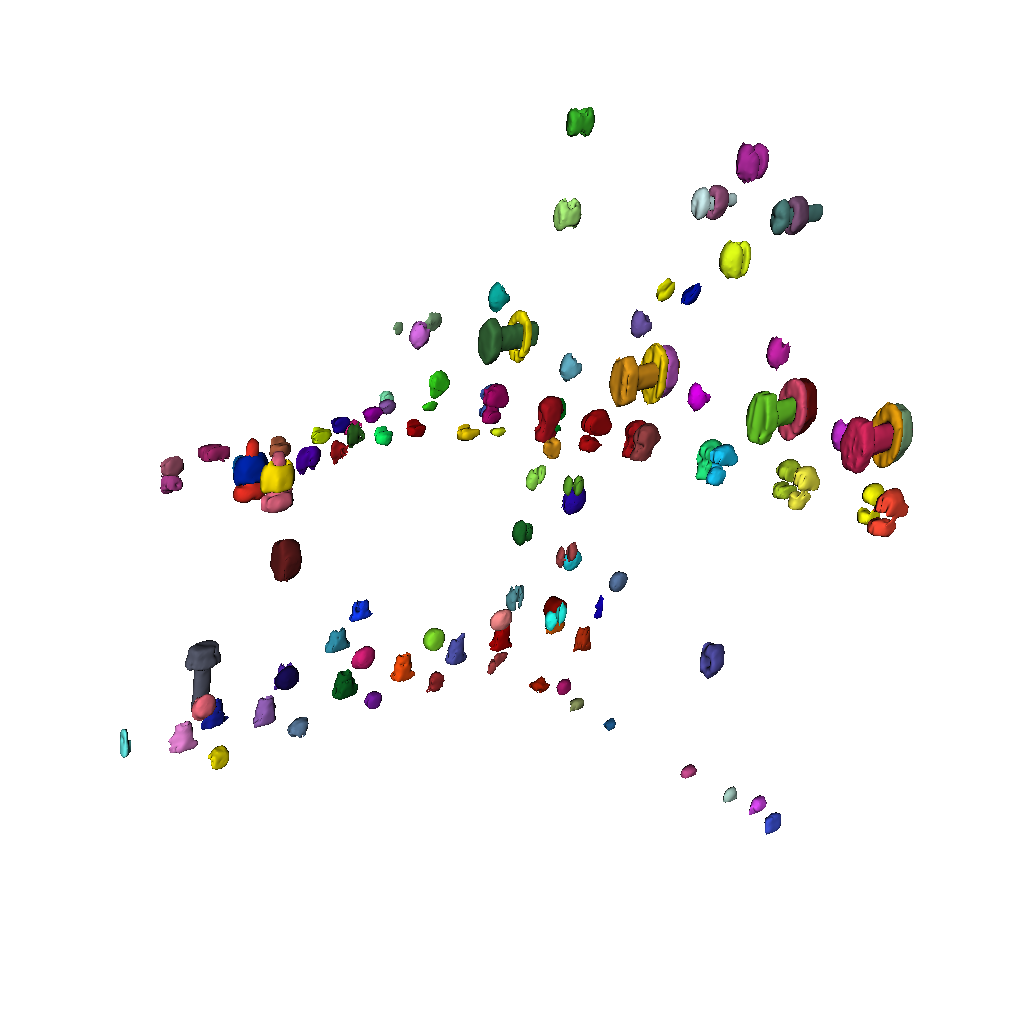}}	
			~\subfloat[\label{fig:description-example-mountingAndMiscellaneous}]
			{\includegraphics[width=0.33\textwidth,height=0.25\textheight,keepaspectratio]
				{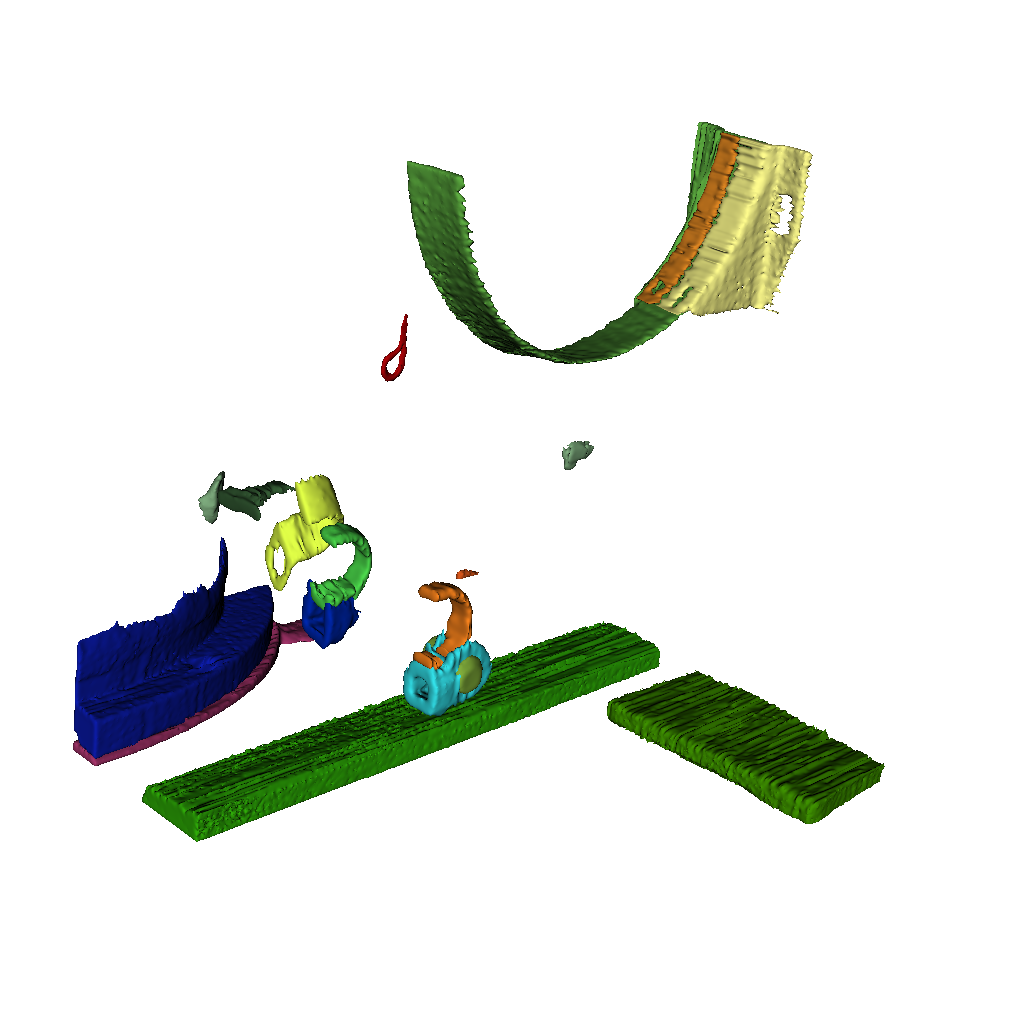}}
		\end{centering}
		\caption{\label{fig:description-example} Example renderings of sub-volume $V_6$ (3072,7168,0). While Figure \ref{fig:description-example-volume} shows the unannotated volume, Figure \ref{fig:description-example-all} depicts all labelled segments separated by colour. To increase clarity, only the segments of a specific category are shown in the following sub-figures: Figure \ref{fig:description-example-metallSheet} provides all metal sheets; Figure \ref{fig:description-example-pipes} gives the presumably pressure-carrying pipes, pressure tanks and lines; Figure \ref{fig:description-example-rivetsAndBolts} contains all rivets and screw connections; Figure \ref{fig:description-example-mountingAndMiscellaneous} finally shows all brackets, clamp connectors and other miscellaneous transition elements that could not otherwise be assigned a category.}
	\end{figure*}

	Figure \ref{fig:description-example} shows different renderings of sub-volume $V_6$ (3072,7168,0). While Figure \ref{fig:description-example-volume} depicts the unannotated volume, Figure \ref{fig:description-example-all} shows all labelled segments separated by colour. To increase clarity, only the segments of a specific category are shown in the following: Figure \ref{fig:description-example-metallSheet} provides all metal sheets; Figure \ref{fig:description-example-pipes} gives the presumably pressure-carrying pipes, pressure tanks and lines; Figure \ref{fig:description-example-rivetsAndBolts} contains all rivets and screw connections; Finally, Figure \ref{fig:description-example-mountingAndMiscellaneous} shows all brackets, clamp connectors and other miscellaneous transition elements that could not otherwise be assigned a category.
	
	\subsection{Annotation pipeline} \label{sec:annotation-pipeline}
	The annotation process of the XXL-CT data is on one side related to the used annotation software (Section \ref{sec:annotation-software}) and annotation hardware (see Section \ref {sec:annotation-hardware}). On the other side, it is highly dependent on the annotation rules and guidelines provided to the annotators as well as institutional knowledge which gets developed over time (see Section \ref{sec:annotation-guidlines}). Furthermore, some postprocessing possibilities such as filtering, morphological operators, or data fusion must be considered (see Section \ref{sec:annotation-post-processing}).
	
	\subsubsection{Annotation Software} \label{sec:annotation-software}						
	We used the application {\em 3D Slicer} \cite{Slicer3D, Fedorov2012} for most of the annotation. This software provides the annotator with different types of interactive annotation tools such as \enquote{paint strokes},  \enquote{boolean operations} or gray value aware \enquote{fill methods} to select individual voxels and voxel groups. Furthermore, it can easily be extended with new segmentation functions \cite{2014-Kikinis-3D} and includes a powerful scripting interface. 
	
	\subsubsection{Annotation Hardware} \label{sec:annotation-hardware}
\begin{figure}
		\centering{}
		\includegraphics[width=1.0\columnwidth,keepaspectratio]{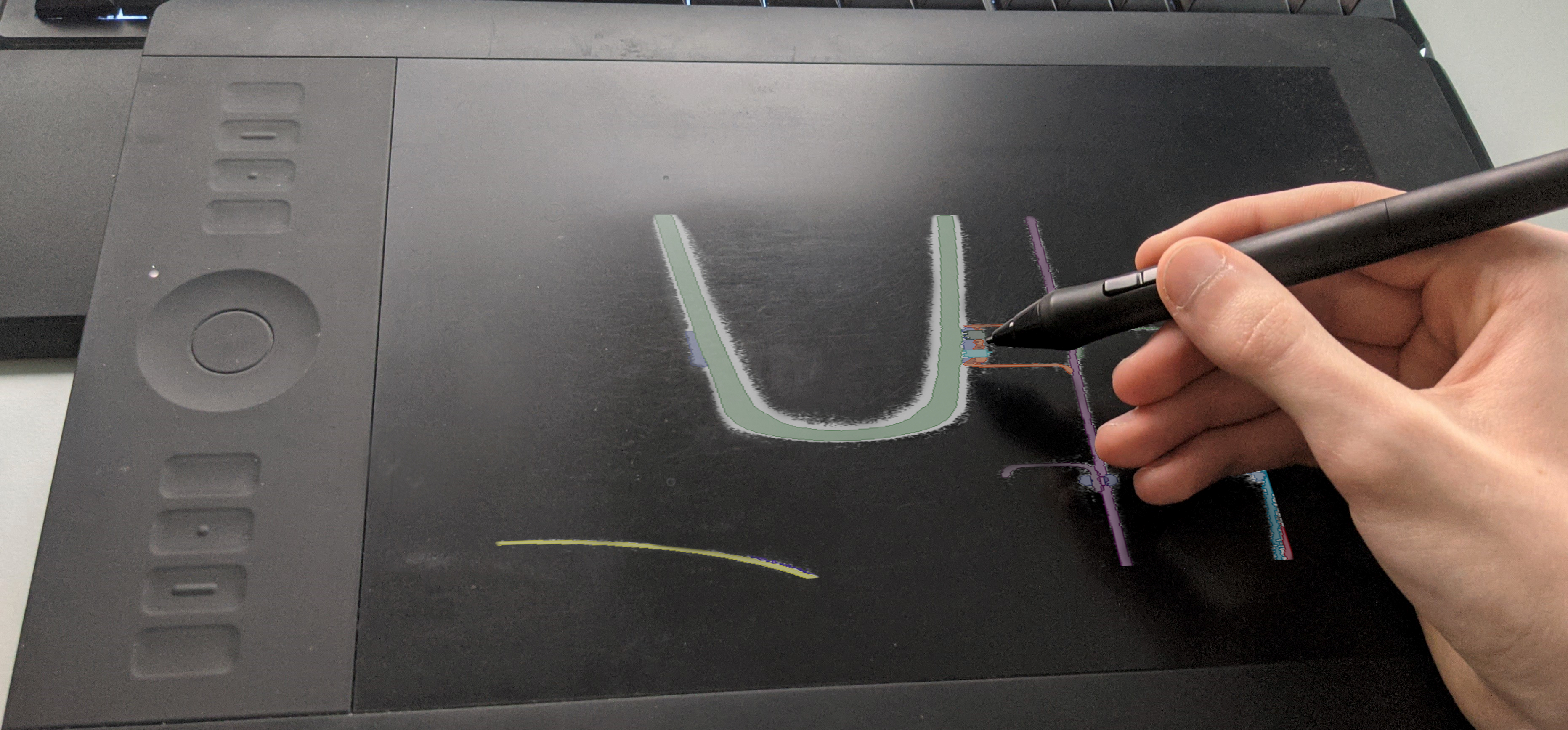}
		\caption{\label{fig:annotation-tablet}Manual labelling of large-scale industrial CT data of an aeroplane part using a high-resolution graphics tablet and a digital pen (Composed image to illustrate the process).}
	\end{figure}

	We used graphic tablets with digital styluses as input devices for the slice-by-slice manual annotation and labelling of the sub-volumes. As they allowed easy and intuitive drawing. In contrast to the use of a mouse this approach is more precise, intuitive and more importantly more gentle on the wrist of the annotators \cite{Dach-2011-DGBMT}. In Figure \ref{fig:annotation-tablet} a typical manual segmentation and labelling task of a sub-volume from XXL data can be seen using a graphics tablet.
	
	\subsubsection{Annotation guidlines} \label{sec:annotation-guidlines}
	In our annotation guidelines, provided to all annotators, we stipulated that the ‘human interpreted reality’ of the data (based on the a-priori knowledge about the depicted objects) and not the ‘perceived visual representation’ should be segmented. For example, if scattered radiation artefacts were encountered, represented through bright or dark streaks through the volume or cupping artefacts from beam hardening, it was suggested to annotate the guessed \emph{real specimen} and not the \emph{distorted image}. 
	
	This should increase the uniformity of the annotations since otherwise it is difficult to find the same thresholds over different volume regions and artifacts. The ultimate goal of the work is to develop methods to separate all components from each other in a meaningful way. This may not be achievable in some cases, e.g. if there is not enough data available. However, this can only be known when everything has been tried, for which a meaningful annotation of the desired ideal result has to be available. 					
	
	\subsubsection{Annotation Post-Processing} \label{sec:annotation-post-processing}
	After the individual segments have been partially annotated automatically by hand, they usually do not yet have the quality expected from a ground truth. Due to the presence of noise on the segment surfaces and voxels that were annotated as belonging to more than one segment, post-treatment is necessary. 
	
\begin{figure}
		\begin{centering}
			\subfloat[\label{fig:annotation-changesThroughPostprocessing-input-XY}]
			{\includegraphics[width=0.33\columnwidth, interpolate=false,height=0.25\textheight,keepaspectratio]
				{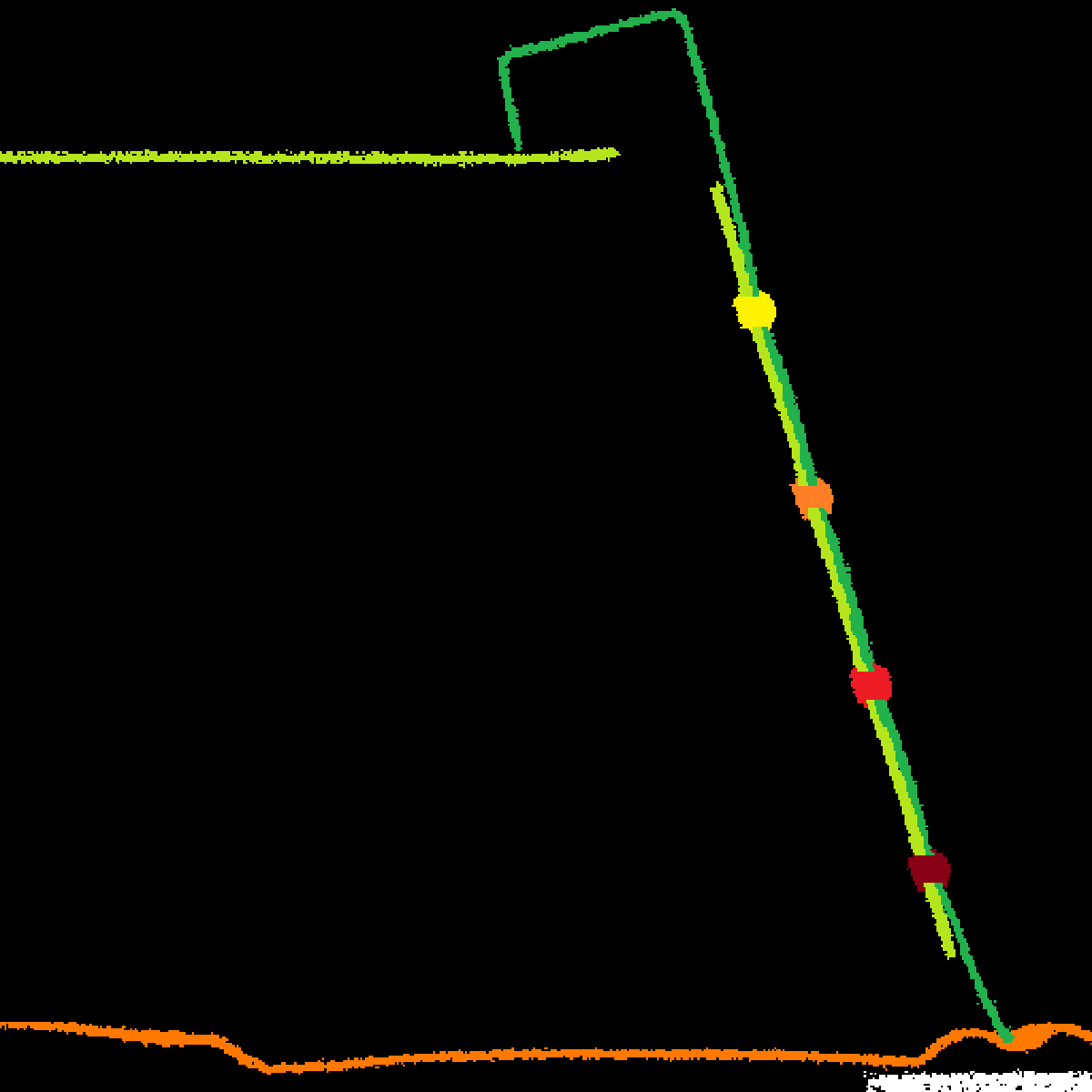}}	
			~\subfloat[\label{fig:annotation-changesThroughPostprocessing-output-XY}]
			{\includegraphics[width=0.33\columnwidth, interpolate=false,height=0.25\textheight,keepaspectratio]
				{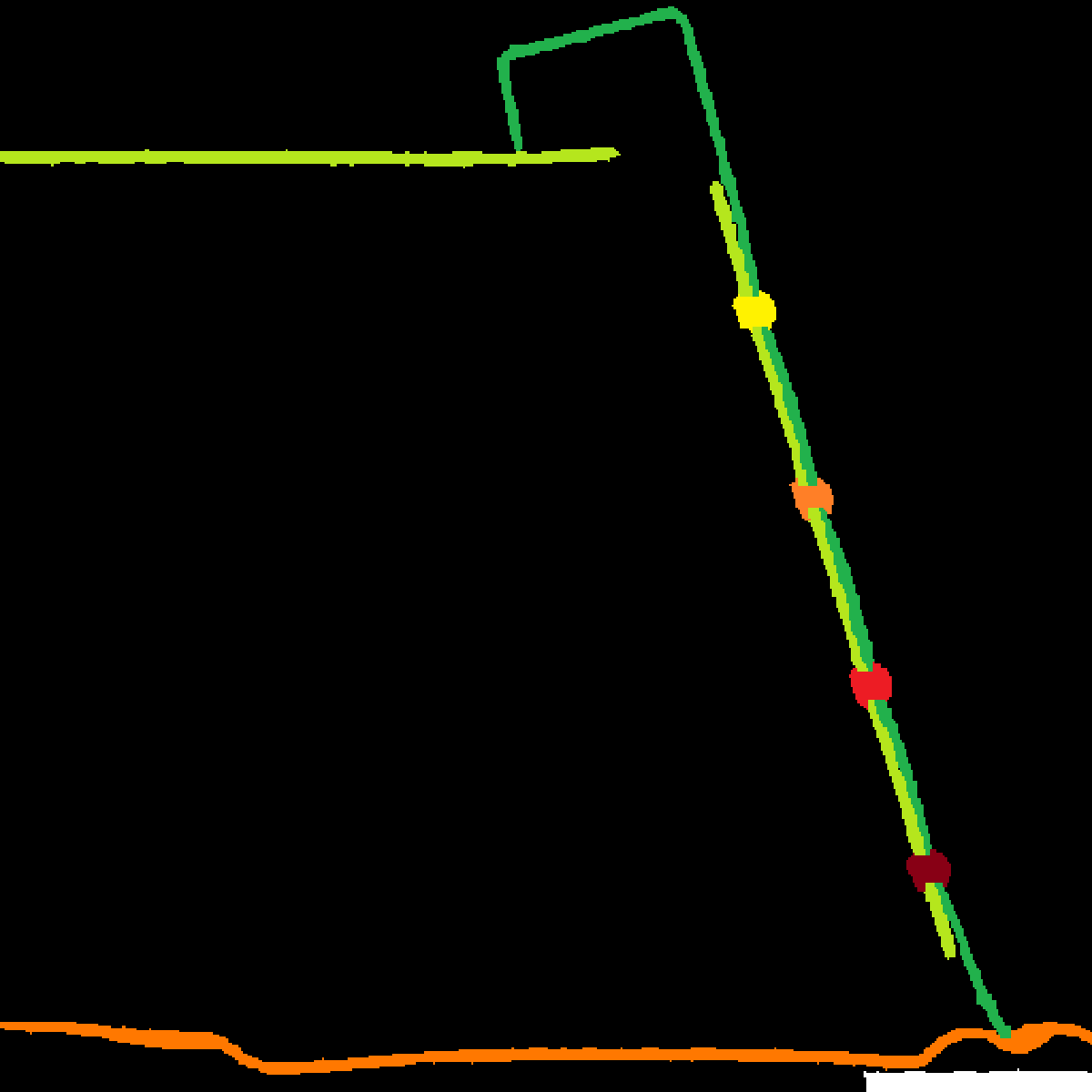}}
			~\subfloat[\label{fig:annotation-changesThroughPostprocessing-difference-XY}]
			{\includegraphics[width=0.33\columnwidth, interpolate=false,height=0.25\textheight,keepaspectratio]
				{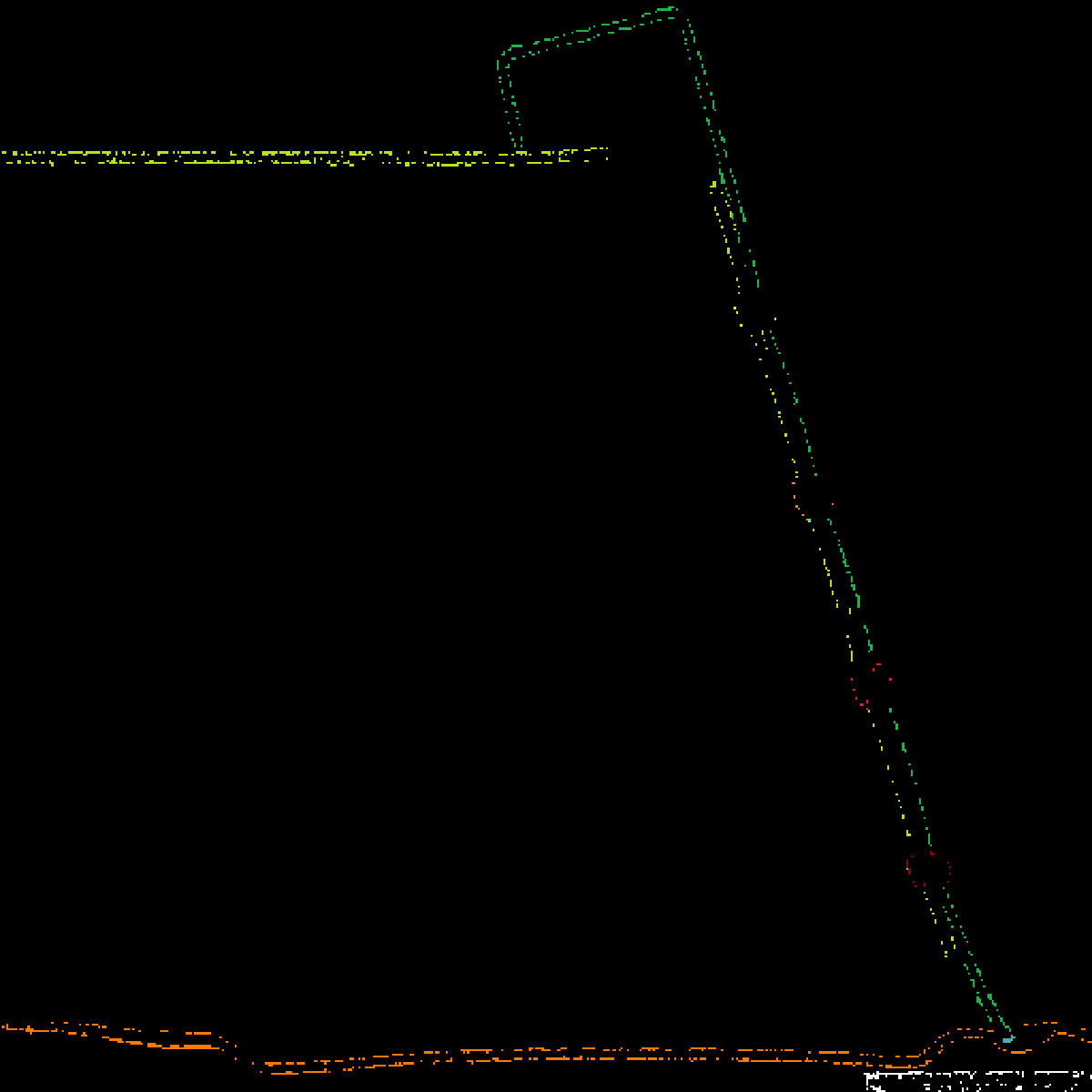}}
			
			\subfloat[\label{fig:annotation-changesThroughPostprocessing-input-XZ}]
			{\includegraphics[width=0.33\columnwidth, interpolate=false,height=0.25\textheight,keepaspectratio]
				{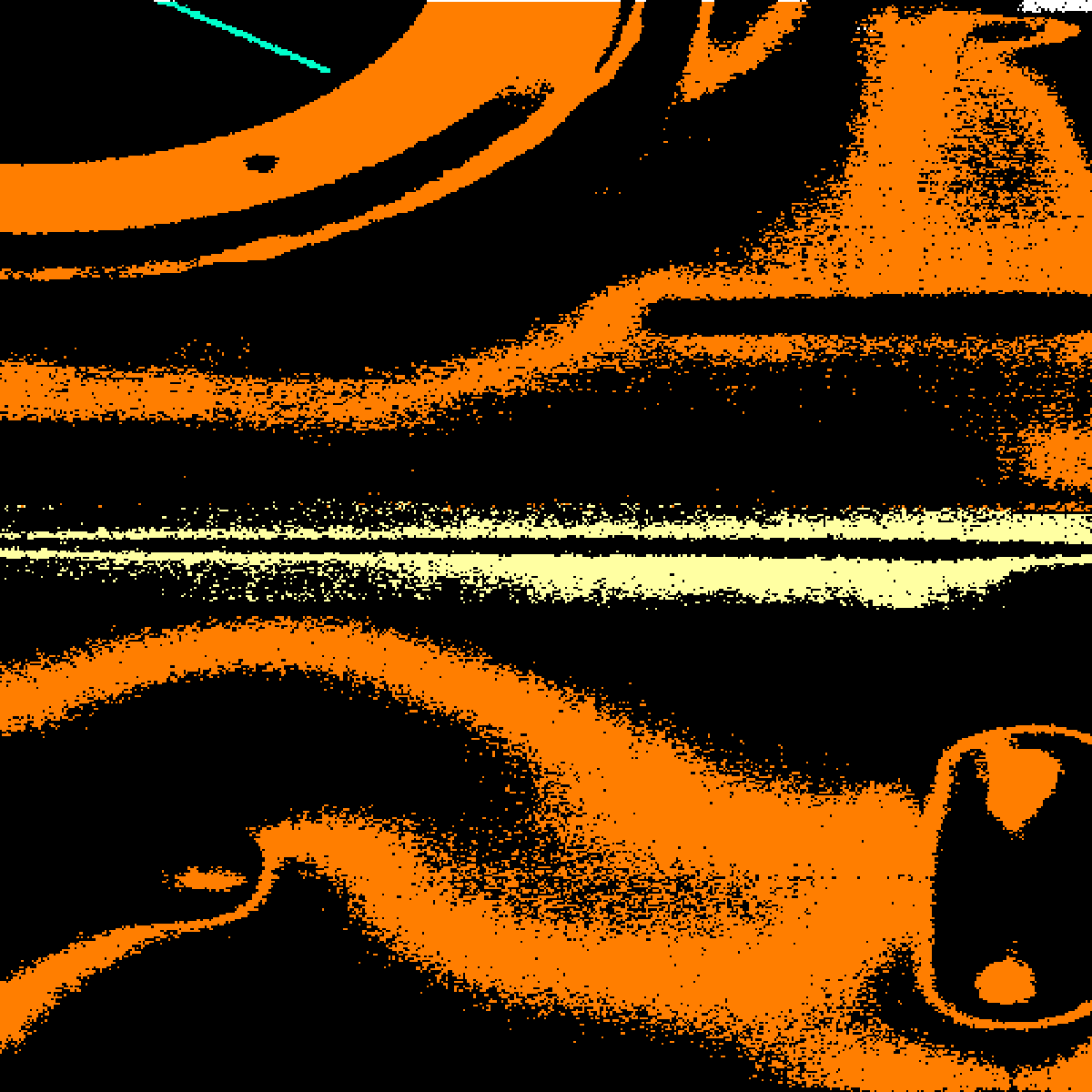}}	
			~\subfloat[\label{fig:annotation-changesThroughPostprocessing-output-XZ}]
			{\includegraphics[width=0.33\columnwidth, interpolate=false,height=0.25\textheight,keepaspectratio]
				{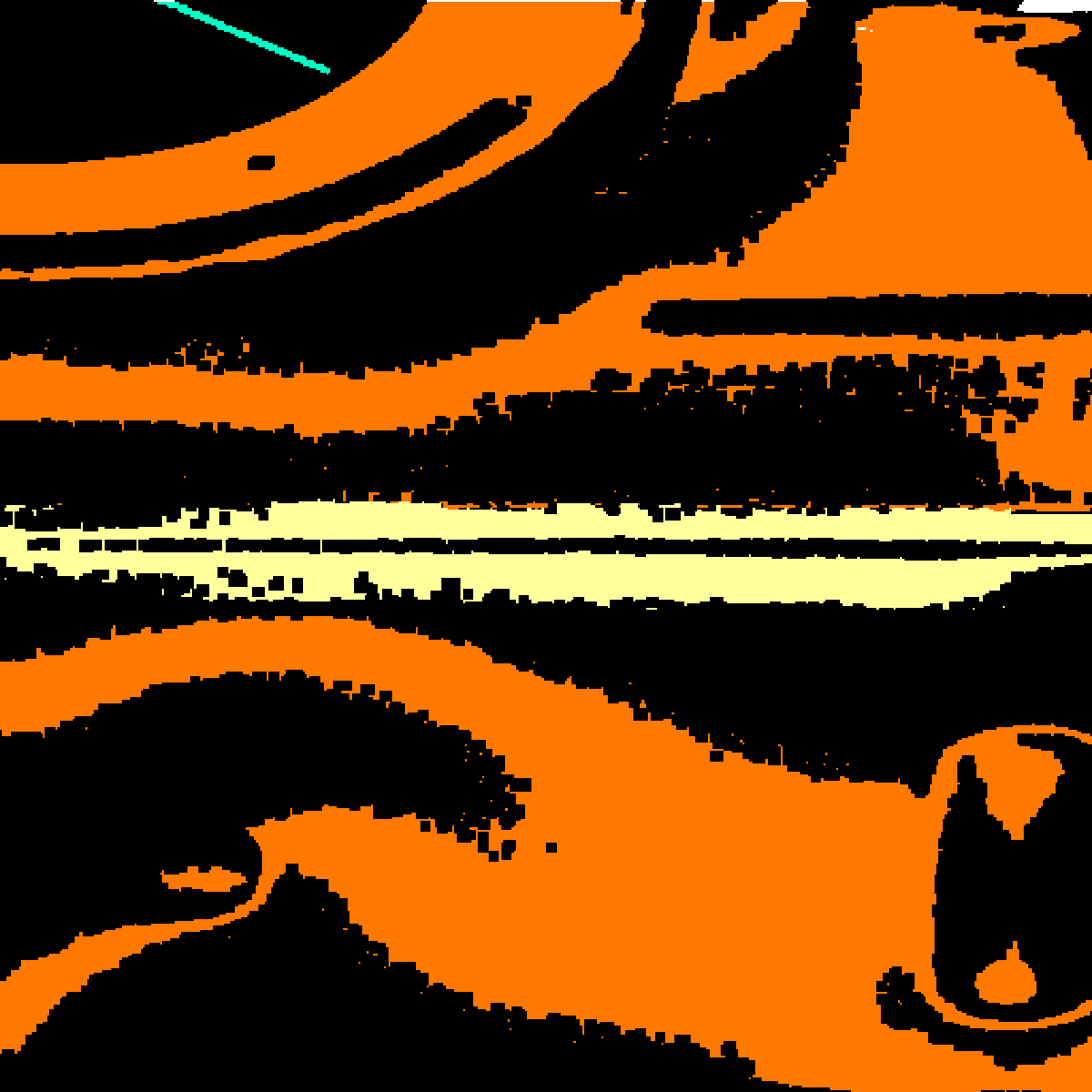}}	
			~\subfloat[\label{fig:annotation-changesThroughPostprocessing-difference-XZ}]
			{\includegraphics[width=0.33\columnwidth, interpolate=false,height=0.25\textheight,keepaspectratio]
				{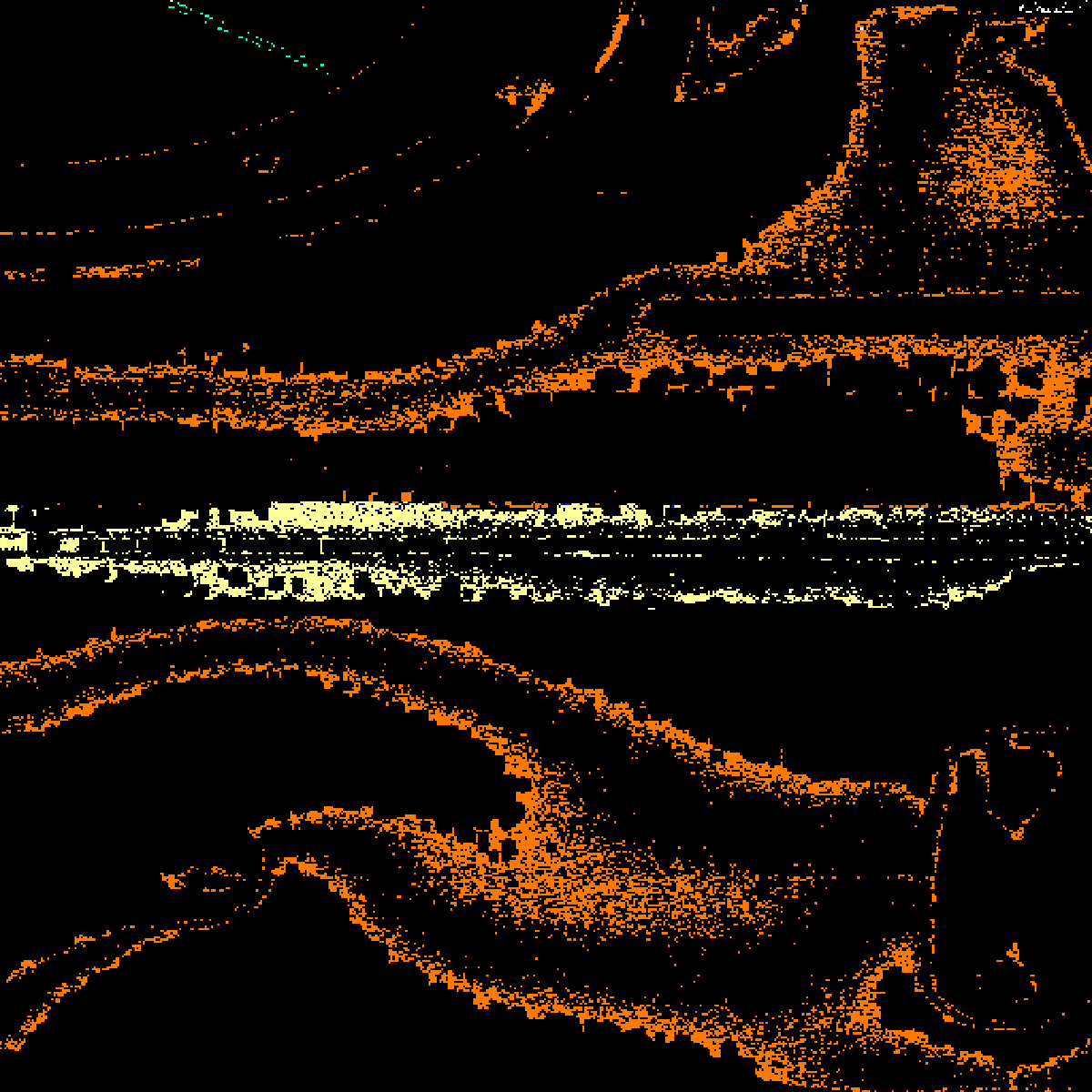}}
		\end{centering}
		\caption{\label{fig:annotation-changesThroughPostprocessing} Slices from sub-volume $V_4$ (3072,6144,0) depicting typical changes introduced by morphological postprocessing. Figures  \ref{fig:annotation-changesThroughPostprocessing-input-XY} and \ref{fig:annotation-changesThroughPostprocessing-input-XZ} (1st column) show manually annotated input volumes. Figures \ref{fig:annotation-changesThroughPostprocessing-output-XY} and \ref{fig:annotation-changesThroughPostprocessing-output-XZ} (2nd column) depict the morphologically postprocessed output. Finally, Figures \ref{fig:annotation-changesThroughPostprocessing-difference-XY} and \ref{fig:annotation-changesThroughPostprocessing-difference-XZ} (3rd column) show the difference between the input and output volumes. The upper row shows an example where the changes introduced by the postprocessing consist mainly of small voxel alterations of the surface of a thin metal sheet. The bottom row depicts the changes close to the surface of the orange metal sheet located at the bottom of the upper row of images. This metal sheet appears to be quite noisy and therefore prone to the more pronounced changes visible in the residual Figure \ref{fig:annotation-changesThroughPostprocessing-difference-XZ}.}
	\end{figure}

	\paragraph{Morphological Closing:}					
	The use of the previously mentioned bandpass filter to visually smooth the grey values sometimes yields grainy textures inside the segments (see example in Section \ref{sec:challenges-noisyData}). Due to the presence of this coarse-grained noise, we decided to postprocess the results obtained by manual annotation to close gaps between the quality of the manual annotation and the desired quality of the segmentation. Overall, it was aimed to achieve semantical reasonable and simultaneously visually pleasing segmentation results. For this purpose, the manual annotation of each segment was first postprocessed using a morphological closing filter \cite{Gonzalez2006} with a $3\times3\times3$ structure element. Figure \ref{fig:annotation-changesThroughPostprocessing} depicts two orthogonal slices from the manually segmented sub-volume $V_4$ (3072,6144,0) prior and post morphological processing. While most of the changes introduced by the postprocessing consist mostly out of simple surface voxel alterations (see Figure \ref{fig:annotation-changesThroughPostprocessing-difference-XY}), they may also include changes to the surfaces of \emph{noisy} metal sheets (see Figure \ref{fig:annotation-changesThroughPostprocessing-difference-XZ}) which are  prone to the more pronounced changes due to their \emph{noisy} nature.					
	
	\paragraph{Overlapping Entities:}
	We annotated each entity in the sub-volume individually slice by slice. In some rare cases, this yielded results, where we annotated voxels as belonging to multiple segments. For example, if the spatial resolution of the reconstructed volume data (with approximately \SI{0.07}{\cubic\milli\metre} per voxel) was not sufficient enough to represent the exact border between two adjacent thin sheets of metal. It was not always possible to represent this cases in an annotation data-set with only voxel resolution. In such cases, the corresponding voxels were annotated as belonging to several segments. 
	
	After finishing the annotation and labelling process of all depicted entities in a sub-volume $V_i$, all these segmented entities were combined on the voxel level into one single volume. Nevertheless, within this step we allowed the possibility to overwrite already existing voxels of previously included segments. The overwriting of labelled voxels primarily occurs at the edges between two adjacent segments. This means that the order in which the segments are processed and fused has partially influenced the result of the final segmentation results. Hence the order of the fusion sequence was assigned pseudo-randomly.
	
	\paragraph{Connected Component Analysis:}
	Finally, we performed a successive connected component analysis with a chessboard metric (aka Chebyshev distance or $L_\infty$ norm)\cite{Gonzalez2006} to find the separated chunks. This also allows for a simple fix of the challenges described in Section \ref{sec:challenges-boomerang}. Furthermore, we discarded small segments with less than 100\,voxels and deleted them, to avoid over-segmentation. The threshold of $\theta = 100$\,voxels was determined empirically. 
	
	\section{Challenges} \label{sec:challenges} 		
	\begin{figure}
		\centering
		\includegraphics[width=0.8\columnwidth,keepaspectratio]{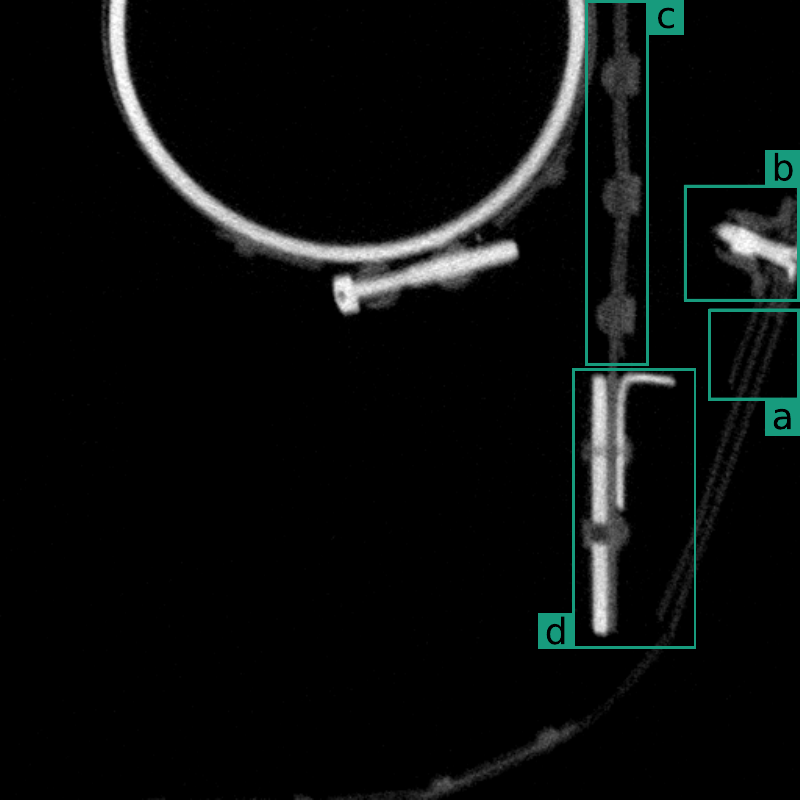}
		\caption{\label{fig:challenges-map} Sectional view from sub-volume {$V_6$} (3072,7168,0) in which some challenging areas have been marked: (a) noisy metal sheets (see Section \ref{sec:challenges-noisyData}); (b) low contrast (see Section \ref{sec:challenges-lowContrast}); (c) low contrast and large contact area (see also Section \ref{sec:challenges-lowContrast}); (d) annotator noise (see Section \ref{sec:challenges-annotator-noise}).}
	\end{figure}		
	
	In the following section, we discuss, some characteristics of the above-introduced data-set and challenges regarding its annotation and labelling. Both, the XXL imaging as well as the labelling steps provide ambiguities with respect to the data. To this end an example from sub-volume $V_6$ (3072,7168,0) will be taken, see Figure \ref{fig:challenges-map}, and used as representative for the corresponding categories of challenges, namely  noise (see Section \ref{sec:challenges-noisyData}), low contrast segments (see Section \ref{sec:challenges-lowContrast}), segments leaving and re-entering the sub-volume (see Section \ref{sec:challenges-boomerang}) as well as annotator noise \ref{sec:challenges-annotator-noise}. However, these categories are only exemplary and not to be understood as fully comprehensive.
	
	\subsection{Noise} \label{sec:challenges-noisyData}
\begin{figure}
		\centering{}
		\subfloat[\label{fig:challenges-noisyData-input}]
		{\includegraphics[width=0.33\columnwidth, interpolate=false,keepaspectratio]
			{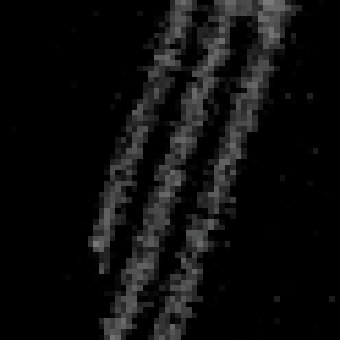}}
		~\subfloat[\label{fig:challenges-noisyData-segmentation}]
		{\includegraphics[width=0.33\columnwidth, interpolate=false,keepaspectratio]
			{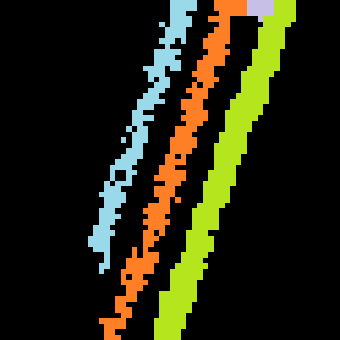}}	
		~\subfloat[\label{fig:challenges-noisyData-closed}]
		{\includegraphics[width=0.33\columnwidth, interpolate=false,keepaspectratio]
			{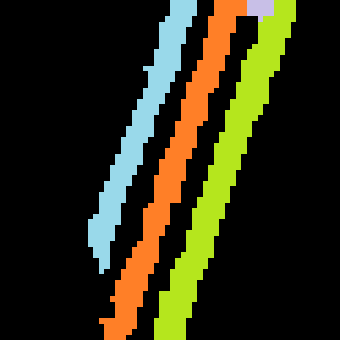}}
		\caption{\label{fig:challenges-noisyData}Three parallel metal plates with high noise in the reconstruction. Figure \ref{fig:challenges-noisyData-input}: enlarged section from sub-volume {$V_6$} (3072,7168,0) (see Figure \ref{fig:challenges-map}~(a)). The grainy texture is due to the low data quality and should therefore not be included in the annotation. Figure \ref{fig:challenges-noisyData-segmentation}: result of naive segmentation; Figure \ref{fig:challenges-noisyData-closed}: desired segmentation after morphological closing.}
	\end{figure}

	Figure \ref{fig:challenges-map} shows at location (a) a region in which three parallel thin metal sheets are visible. In Figure \ref{fig:challenges-noisyData-input} an enlarged version is depicted where it can be observed that the three metal plates are interspersed with coarse-grained noise. Figure \ref{fig:challenges-noisyData-segmentation} provides the naive annotation strictly based on the visible grey values, leading to a result permeated by granular noise. However, using a-priori knowledge that the displayed metal components do not consist of sponge-like porous material, but the coarse-grained texture is due to measurement or reconstruction artefacts, the annotation is modified using the morphological closing (see above) as postprocessing step, yielding the desired result shown in Figure \ref{fig:challenges-noisyData-closed}. 				
	
	\subsection{Low to no contrast between segments} \label{sec:challenges-lowContrast}			
\begin{figure}
		\centering{}
		\subfloat[\label{fig:challenges-lowContrast-input}]
		{
			\begin{tikzpicture}
				\node [anchor=south west,inner sep=0] (image) at (0,0) {\includegraphics[width=0.31\columnwidth, interpolate=false,keepaspectratio]{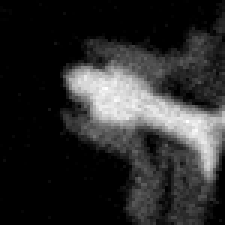}};
				\begin{scope}[x={(image.south east)},y={(image.north west)}]
					\begin{scope}[shift={(0,1)},x={(1/225,0)},y={(0,-1/225)}]
						\draw [thick, dashed, fhg] (84,155) -- (144,35);
					\end{scope}
				\end{scope}
			\end{tikzpicture}	
		}
		~\subfloat[\label{fig:challenges-lowContrast-groundTruth}]
		{\includegraphics[width=0.31\columnwidth, interpolate=falsekeepaspectratio]
			{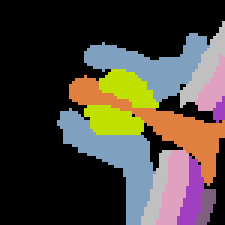}}					
		~\subfloat[\label{fig:challenges-lowContrast-graph}]
		{\includegraphics[width=0.35\columnwidth, keepaspectratio]
			{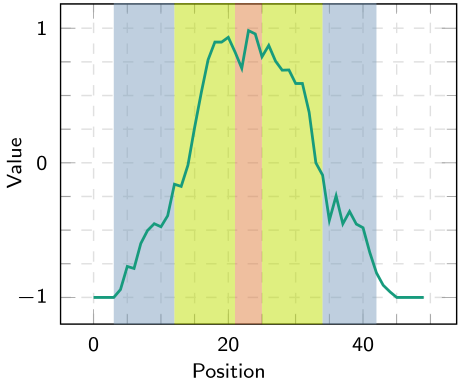}}
		\caption{\label{fig:challenges-lowContrast} Example of low to no contrast entities. Figure \ref{fig:challenges-lowContrast-input}: Slice from sub-volume {$V_6$} (3072,7168,0) (see Figure \ref{fig:challenges-map}~(b)) presumably showing a screw and its corresponding nut. No appreciable grey value and texture differences between the two components can be determined. Figure \ref{fig:challenges-lowContrast-groundTruth}: possible semantic annotation with an orange screw (\tikzBoxOrange) and a light green nut (\tikzBoxLightGreen) inside a blue structure (\tikzBoxBlue); Figure \ref{fig:challenges-lowContrast-graph}: grey value profile plot along the green dashed section marked in the left subfigure, where the background colours indicate the possible annotation into the semantic segments.}
	\end{figure}

	Figure \ref{fig:challenges-map} at location (b) as well as the zoomed-in area in Figure \ref{fig:challenges-lowContrast-input} shows a region in which specifically the bright object components to be annotated have no appreciable grey value or texture contrast to each other. Figure \ref{fig:challenges-lowContrast-groundTruth} shows a possible annotation in which the presumed bolt or screw (depicted in orange), runs through the nut (in light green). Figure \ref{fig:challenges-lowContrast-graph} provides the grey value plot on along of the green dashed line in Figure \ref{fig:challenges-lowContrast-input}. The coloured backgrounds refer to the annotation, see Figure \ref{fig:challenges-lowContrast-groundTruth}. This annotation cannot be justified by the existing grey values and textures alone but must be made by examining the neighbouring similar structures and knowledge or assumptions about the production process.
	
\begin{figure}
		\centering{}
		\subfloat[\label{fig:challenges-fullContact-input}]
		{
			\begin{tikzpicture}
				\node [anchor=south west,inner sep=0] (image) at (0,0) {\includegraphics[width=0.3\columnwidth,height=0.25\textheight, interpolate=false,keepaspectratio]{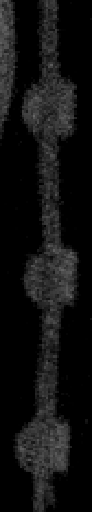}};
				\begin{scope}[x={(image.south east)},y={(image.north west)}]
					\begin{scope}[shift={(0,1)},x={(1/92,0)},y={(0,-1/512)}]
						\draw [thick, dashed, fhg] (25,330) -- (75,330);
					\end{scope}
				\end{scope}
			\end{tikzpicture}	
		}
		~\subfloat[\label{fig:challenges-fullContact-groundTruth}]
		{\includegraphics[width=0.3\columnwidth,height=0.25\textheight, interpolate=false,keepaspectratio]
			{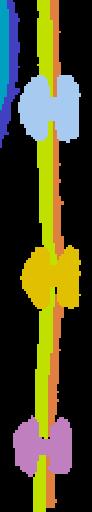}}	
		~\subfloat[\label{fig:challenges-fullContact-graph}]
		{\includegraphics[width=0.35\columnwidth, keepaspectratio]
			{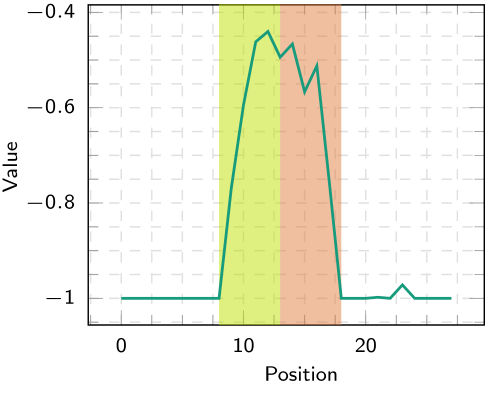}}			
		\caption{\label{fig:challenges-fullContact} Example of low to no contrast entities.   Figure \ref{fig:challenges-fullContact-input} shows a slice from sub-volume $V_6$ (2072,7168,0) (see Figure \ref{fig:challenges-map}~(c)) containing two metal sheets riveted together. No appreciable grey value and texture differences between the two components can be determined.   Figure \ref{fig:challenges-fullContact-groundTruth}: possible manual annotation of the left (light green (\tikzBoxLightGreen)) and right (orange (\tikzBoxOrange)) metal sheets.  Figure \ref{fig:challenges-fullContact-graph}: grey value plot along the green dashed section marked in   Figure \ref{fig:challenges-fullContact-input}. The background colors indicate the possible annotation into semantic segments.
		}
	\end{figure}

	Another example of such low-contrast segment boundaries between adjacent metal sheets is shown in the field of view in Figure \ref{fig:challenges-map}~(c) or in Figure \ref{fig:challenges-fullContact}. Figure \ref{fig:challenges-fullContact-groundTruth} depicts a possible manual segmentation of the two metal sheets and the rivets in the regions. Figure \ref{fig:challenges-fullContact-graph} shows a grey value profile of the green dashed line shown in Figure \ref{fig:challenges-fullContact-groundTruth}, together with its possible segmentation as a colored background. Similar to before, the course of the segment boundaries can only be argued using a-priori knowledge from the surrounding segments and layers.				
	
\begin{figure}
		\centering{}
		\subfloat[\label{fig:challenges-rivet-input}]
		{\includegraphics[width=0.33\columnwidth,height=0.25\textheight, interpolate=false,keepaspectratio]
			{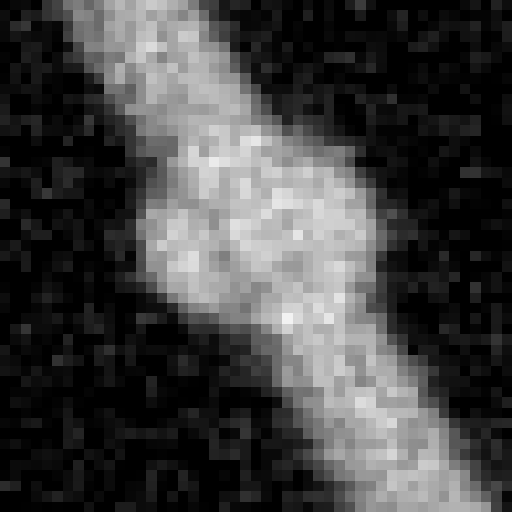}}
		~\subfloat[\label{fig:challenges-rivet-groundTruth}]
		{\includegraphics[width=0.33\columnwidth,height=0.25\textheight, interpolate=false,keepaspectratio]
			{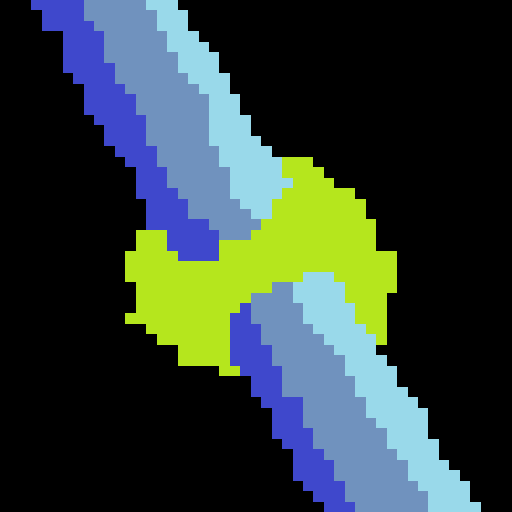}}	
		\caption{\label{fig:challenges-rivet} Example of an entity with low to no contrast.  Figure \ref{fig:challenges-rivet-input}: slice from sub-volume $V_3$ (3072,5632,0) presumably depicting three metal sheets riveted together. No appreciable visual grey values nor texture differences between the components can be determined. Figure \ref{fig:challenges-rivet-groundTruth}: possible (assumed) annotation of the regions taking the surrounding topology into account.} 
	\end{figure}

	Finally, Figure \ref{fig:challenges-rivet} shows a similar case from sub-volume {$V_3$}. Here, a rivet penetrates three adjacent metal sheets. Due to similar material densities and the large and evenly shaped contact surface, the transition between the rivet and sheet metal cannot be discerned clearly. 
	
	\subsection{Re-entering segments} \label{sec:challenges-boomerang}						
\begin{figure}
		\centering{}
		\subfloat[\label{fig:challenges-boomerang-map}]
		{\includegraphics[height=0.25\textheight, interpolate=false,keepaspectratio]{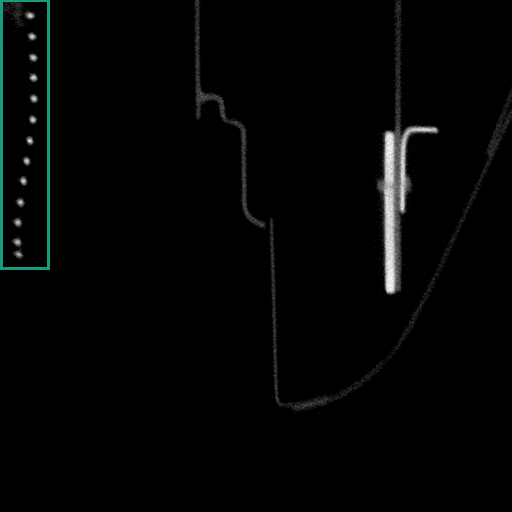}}
		~
		\subfloat[\label{fig:challenges-boomerang-input}]
		{\includegraphics[height=0.25\textheight, interpolate=false,keepaspectratio]{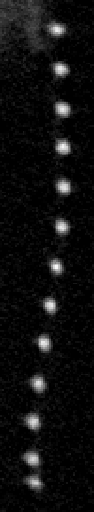}}
		~
		\subfloat[\label{fig:challenges-boomerang-segmentation-uncorrected}]
		{\includegraphics[height=0.25\textheight, interpolate=false,keepaspectratio]{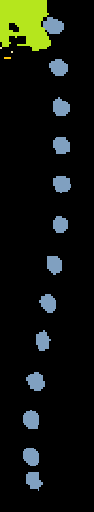}}
		~
		\subfloat[\label{fig:challenges-boomerang-segmentation-corrected}]
		{\includegraphics[height=0.25\textheight, interpolate=false,keepaspectratio]{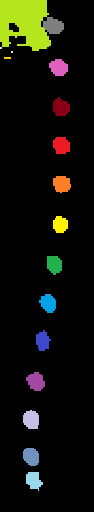}}
		\caption{\label{fig:challenges-boomerang} Example of a slice (from sub-volume $V_6$ (3072,7168,0)\,) depicting a component which is not fully contained in the current sub-volume. Figure \ref{fig:challenges-boomerang-map} and \ref{fig:challenges-boomerang-input}: helical wire support structure. Without further information, the individual coils appear to be thirteen separate segments; Figure \ref{fig:challenges-boomerang-segmentation-uncorrected}: the result of human segmentation; Figure \ref{fig:challenges-boomerang-segmentation-corrected}: annotation result after connected component analysis, where no correspondences among the entities have been found.}
	\end{figure}

	Some components in the volumetric data leave the visible area of the current sub-volume and reappear as disconnected segments at a different location of the same sub-volume (see Figure \ref{fig:challenges-boomerang}). Here the component of interest --- a helical wire support structure probably for a suction hose --- is located in the upper left corner of a sub-volume see Figure \ref{fig:challenges-boomerang-map} for overview and Figure \ref{fig:challenges-boomerang-input} for an enlarged view. Without any further semantic information, the individual coils appear to be thirteen separate segments. Figure \ref{fig:challenges-boomerang-segmentation-uncorrected} depicts the result of a human segmentation of these entities. Figure \ref{fig:challenges-boomerang-segmentation-corrected} provides the final annotation result after applying a connected component analysis, where no correspondences and connections among the thirteen entities have been found.
	
	However, without additional semantic information about the course of the entities outside the sub-volume, it must be assumed that these segments are most likely separated from each other. For this reason, we performed a connected component analysis on the hand-annotated data-set and separated these segments as they leave and re-enter the sub-volume.
	
	\subsection{Annotator noise} \label{sec:challenges-annotator-noise}
	Limited knowledge of the true real ground truth often leads to severe annotator noise \cite{x58, Vadineanu2022}, which can often be observed within vast and difficult-to-label data sets. Different annotators will have inconsistent knowledge of the problem domain, are possibly fatigued, subconsciously introduce their own bias into the annotation output, or will label multiple parts differently. Thus, the obtained annotation from a specific annotator or a fusion of several annotations should be only understood as one possible annotation.
	
\begin{figure}
		\centering{}
		\subfloat[\label{fig:challenges-annotatorNoise-input}]
		{\includegraphics[width=0.23\columnwidth, interpolate=false,keepaspectratio]{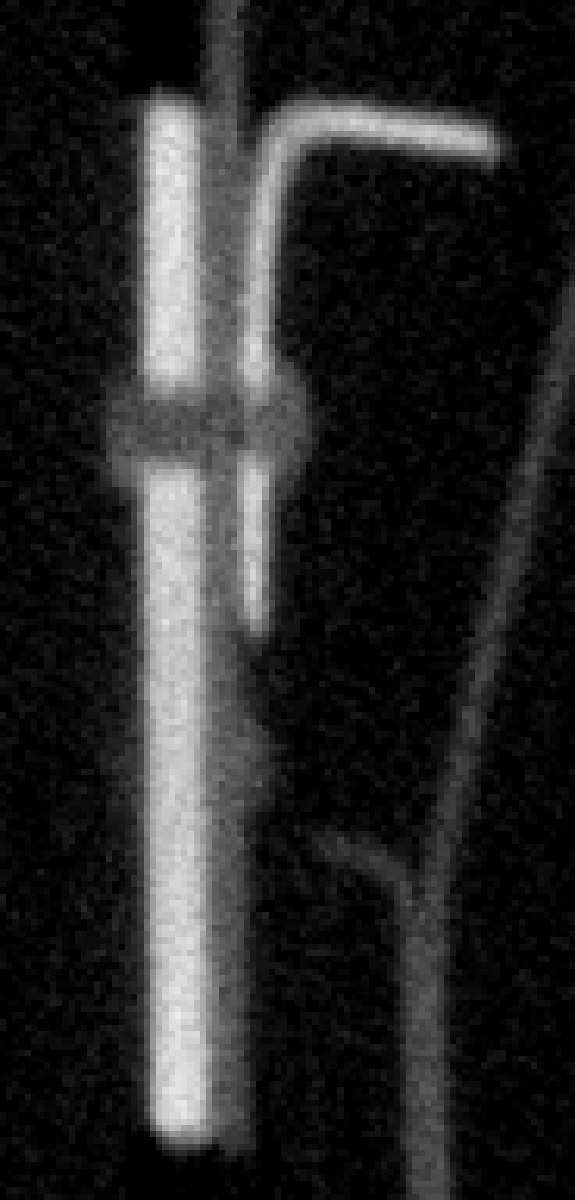}}
		~
		\subfloat[\label{fig:challenges-annotatorNoise-first}]
		{\includegraphics[width=0.23\columnwidth, interpolate=false,keepaspectratio]{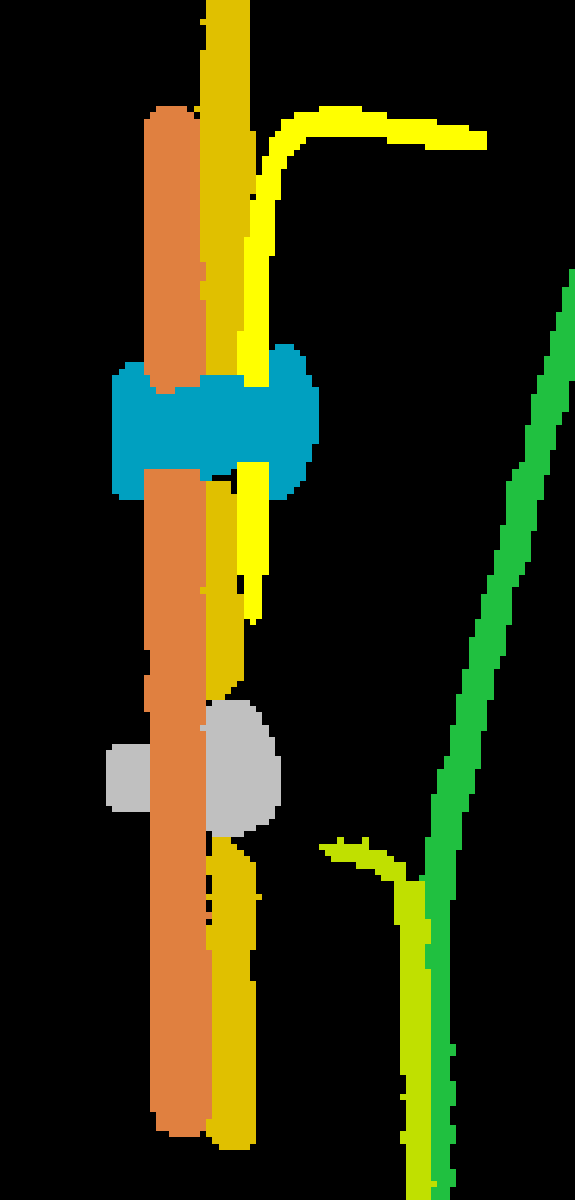}}
		~
		\subfloat[\label{fig:challenges-annotatorNoise-second}]
		{\includegraphics[width=0.23\columnwidth, interpolate=false,keepaspectratio]{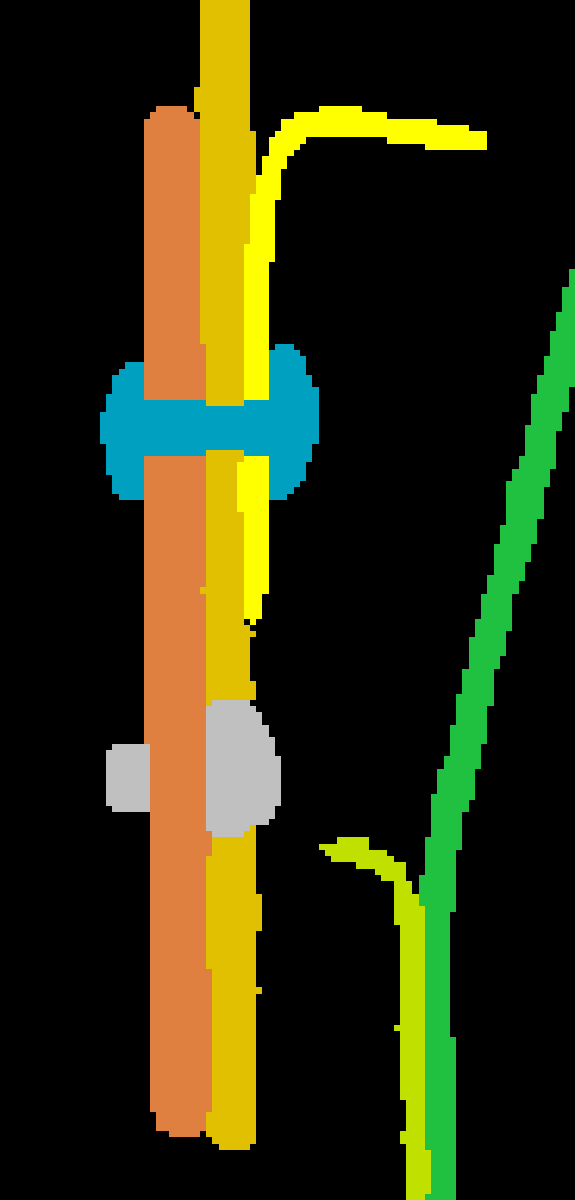}}
		~
		\subfloat[\label{fig:challenges-annotatorNoise-difference}]
		{\includegraphics[width=0.23\columnwidth, interpolate=false,keepaspectratio]{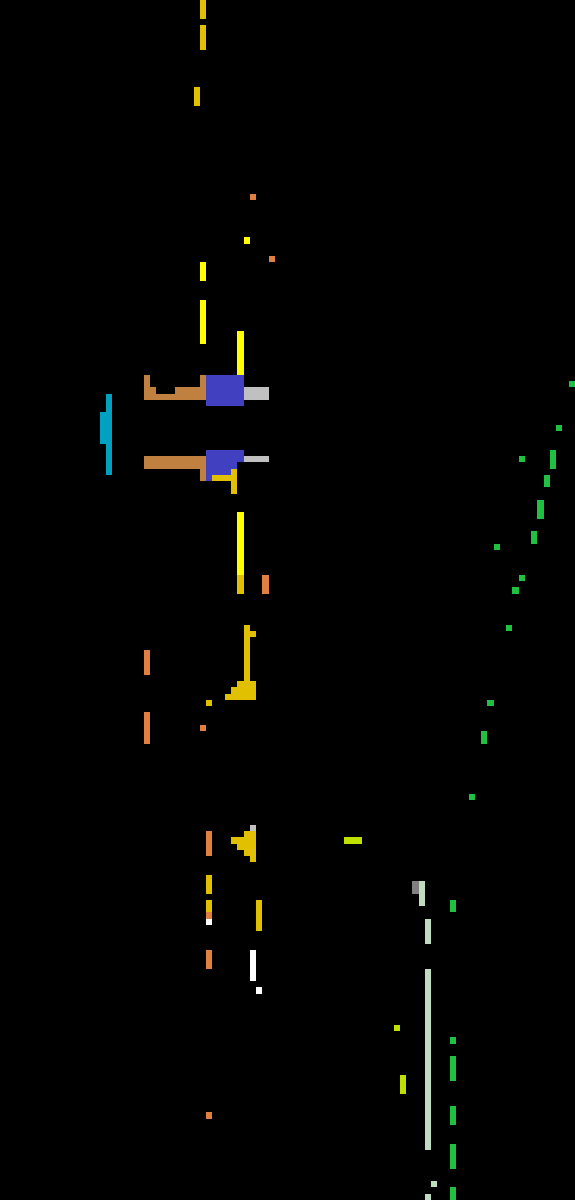}}
		
		\caption{\label{fig:challenges-annotatorNoise} Example of a small region (from Figure \ref{fig:challenges-annotatorNoise-input}~(d)) depicting multiple metal sheets riveted together annotated by two different annotators (Figures \ref{fig:challenges-annotatorNoise-first} and \ref{fig:challenges-annotatorNoise-second}). The differences between both annotators are shown in   Figure \ref{fig:challenges-annotatorNoise-difference}.}
	\end{figure}

	Figure \ref{fig:challenges-annotatorNoise} shows a small region (Figure \ref{fig:challenges-annotatorNoise-input} (d)) of sub-volume {$V_6$} (3072,7168,0) which has been annotated by two different annotators (see Figures \ref{fig:challenges-annotatorNoise-first} and \ref{fig:challenges-annotatorNoise-second}). Additionally, the difference volume between the two annotations is depicted in Figure \ref {fig:challenges-annotatorNoise}. As can be seen, most of the metal sheets only diverge in some surface voxels, whereas the rivet was annotated quite differently by each annotator.
	
	\section{Comparison Metric}	\label{sec:metric}
\begin{figure}
		\centering{}
		{\includegraphics[width=0.5\columnwidth, keepaspectratio]		{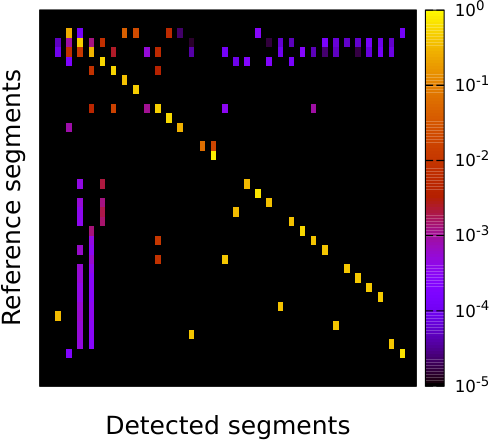}}
		\caption{\label{fig:metric-segmentationComperator-firstSecondAnnotator6144}Correlation matrix of the segmentation of sub-volume $V_4$ (3072,6144,0) of the segmentation results for the same data-set annotated by two different annotators. The rows correspond to \emph{reference segments}, here the first initial annotation. The columns correspond to \emph{detected segments}, here the second refined annotation, which are sorted by the maximum IoU to a \emph{reference segment}.}		
	\end{figure}

	The example of a \emph{segment correlation matrix} depicted in Figure \ref{fig:metric-segmentationComperator-firstSecondAnnotator6144} shows how well the results of two different segmentations provided by two different annotators may match. In this case the \emph{set of reference segments} $S_R$ was initially generated by one annotator and is depicted on the vertical axis of the matrix in Figure \ref{fig:metric-segmentationComperator-firstSecondAnnotator6144}. The \emph{set of detected segments} $S_D$ was created by a second annotator, refining the first segmentation with our current understanding of the data-set. This set is depicted on the horizontal axis of the matrix in Figure \ref{fig:metric-segmentationComperator-firstSecondAnnotator6144}.
	
	Each row is assigned to one \emph{reference segment} $S_{R}(i)$ and each column is assigned to a \emph{detected segment} $S_{D}(j)$. The value or colour of each cell corresponds to the \emph{Intersection over Union} ({\em IoU}) score (also known as {\em Jaccard-Index}) of two segments $S_{R}(i)$ and $S_{D}(j)$:
	
	\[IoU = \frac{|S_{R}(i) \bigcap S_{D}(j)|}{|S_{R}(i) \bigcup S_{D}(j)|}\].	
	
	If these two segments yield a complete overlap (meaning that their segmentations match completely) the value $IoU$ is equal to 1.0. If two compared segments do not share at least one common voxel, the value IoU will be 0.0. All other overlap scenarios are mapped to a value range of $IoU \in [ 0, 1] $.
	
	The rows in the matrix are sorted in descending order by the count of voxels of their corresponding reference segments. Consequently, the top rows correspond to the largest segments and the bottom rows to the smallest segments. The columns have been sorted by searching for the detected segment with the best match, or highest IoU to the reference segment of the current row. Each detected segment can only be assigned to a single reference segment. Detected segments unmatched to a reference segment are sorted by their voxel count. We excluded segments with a voxel count of fewer than 100 voxels to reduce the size of the matrix.
	
	Hence, a {\em perfect segmentation} $S$ with respect to a reference segmentation $R$ should be reflected by a quadratic correlation matrix containing the same count of rows and columns, and thus the same amount of reference segments and detected segments. Additionally, all correlation values outside the main diagonal should contain IoU values of $IoU = 0.0$, while all values on the main diagonal should have values of $IoU = 1.0$.
	
	However, in realistic application examples,	the row and column count will differ. Usually, an over-segmentation will result in more columns than rows. Boundary errors will result in suboptimal correlation values. Rows with multiple horizontal values either denote an over-segmentation of the respective detected segment, or a reference segment that was accidentally been split into multiple segments. In contrast, vertical lines indicate segments spanning multiple reference segments. They merge multiple reference segments. Breaks in the diagonal line indicate reference segments without a good match in the detected segments.
	
\begin{figure}
		\centering{}
		{\includegraphics[width=0.5\columnwidth, keepaspectratio]		{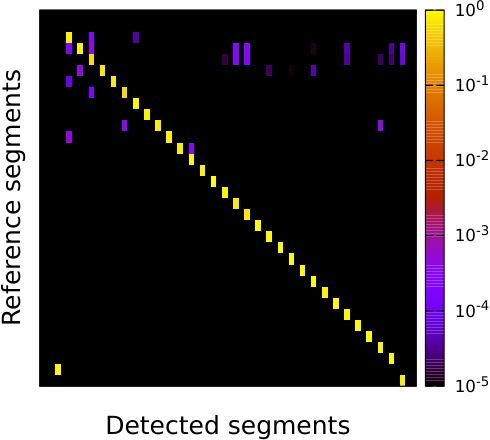}}
		\caption{\label{fig:metric-segmentationComperator-priorPosPreparation6144}Correlation matrix of the segmentation of sub-volume $V_4$ (3072,6144,0) before and after the postprocessing. Rows correspond to \emph{reference segments}, here the manual annotation (see Figure \ref{fig:challenges-noisyData-segmentation}), which are sorted top to bottom by decreasing voxel count of the segments. The columns correspond to the \emph{detected segments}, here the postprocessed segments (see Figure \ref{fig:challenges-noisyData-closed}), which are sorted by the maximum IoU to a \emph{reference segment}.}		
	\end{figure}

	Figure \ref{fig:metric-segmentationComperator-priorPosPreparation6144} shows an example result of the manual annotation of a sub-volume compared to the postprocessed version of the same sub-volume. The desired bright diagonal line from the top left to the bottom right is pronounced, indicating that most of the \emph{reference segments} (prior postprocessing) could be assigned to the \emph{detected segments} (after postprocessing). The scattered purple cells, mostly located in the top third of the matrix, signal that some voxels of the manually augmented segments overlap multiple postprocessed segments and are assigned to them. This often happens if the surface of the manual segmentation which was created using the bandpass selection gets smoothed by the postprocessing (see Section \ref{sec:annotation-post-processing}).
	
	Figure \ref{fig:metric-segmentationComperator-firstSecondAnnotator6144} shows the correlation matrix between the two manually annotated version of sub-volume $V_4$ (3072,6144,0) which have been annotated by two different annotators. It can be seen that the annotated segments of the two annotators, especially the smaller segments, mostly match. As the two more or less pronounced vertical lines in towards the left matrix size indicate, most segments annotated by the first annotator lose voxels, most likely surface voxels, to the bigger segments segmented by the second annotator. The gap in the diagonal line almost at the center of the matrix corresponds to a rivet which was annotated much sturdier in the first annotation than by the second annotator.
	
	\section{Discussion and Conclusions}\label{sec:conclusions}
	In this work, we presented a data collection of seven manually-annotated sub-volumes obtained from an XXL-CT Dataset from a historical airplane. These sub-volumes can potentially serve as a novel benchmark date-collection for instance segmentation in the field of non-destructive testing using XXL-CT sub-volumes. To our knowledge, at this point of time similar public data sets from XXL-CT are not available.
	
	For the complete XXL-CT volume data we described the acquisition and measurement procedures, as well as its further processing. We described how and according to which criteria the seven sub-volumes were annotated and labelled manually by various annotators, including the description and discussion of challenges regarding possible ambiguities contained in the data-set.
	
	We would like to note that although we have taken great care to annotate the sub-volumes to the best of our knowledge and belief, we may still have made mistakes. Some regions of the data-set simply cannot be clearly annotated due to the quality of the data and the recording modality.
	
	All reconstruction and labelled sub-volumes are available under \cite{fraunhoferMe163InstanceSegmentationDataset}
	
	We hope that the provided data sets are useful for further research.
	
	\subsection*{Acknowledgment}
		This work was supported by the Bavarian Ministry of Economic Affairs, Regional Development and Energy through the Center for Analytics Data Applications (ADA-Center) within the framework of ,,BAYERN DIGITAL II`` (20-3410-2-9-8).
	
		We want to thank the following colleagues for the manual annotation: Verena Malowaniec, Laura Heidner and Kseniia Dudchenko.
	
	\subsection*{Author Contributions}  
		\begin{description}
			\item [{RG}] – organized, supervised und participated in the interactive annotation of the XXL data, he also organized the annotation tools, defined, and built all necessary interfaces and integrated it in the segmentation workflow. He drafted and wrote the manuscript including the graphics.
			\item [{NR}] – together with MB, TF and MS – organized and performed the XXL scan of the aero plane and provided the background information in the paper about the scanning process and setups as well as the scanning parameter.
			\item [{AH}] – together with NS – organized the museums logistics of the airplane scanning and provided the historical background and contextual setting in the paper.
			\item [{SG}] – was involved in the idea and conception of the XXL-data preparation and annotation, as well as in the planning, conception and proofreading of the paper.
			\item [{MB}] – together with NR, TF and MS – organized and performed the XXL scan of the aero plane.
			\item [{TF}] – together with NR, MB and MS – organized and performed the XXL scan of the aero plane.
			\item [{MS}] – together with NR, MB and TF organized, performed, and supervised the XXL scan of the aero plane 
			\item [{TW}] - together with RG and SG – provided the idea, planned, and concepted the paper, wrote the introduction and the setting, and did the final proofreading and editing.
		\end{description}
	
		All authors reviewed the manuscript.
	
	\section*{Competing interests} 
	The authors declare that they have no competing financial and/or non-financial interests in relation to the work described.		
	
	\bibliographystyle{plainurl}
	\bibliography{dataset}
\end{document}